\pdfoutput=1

\documentclass[11pt]{article}

\usepackage{ACL2023}

\usepackage{times}
\usepackage{latexsym}

\usepackage[T1]{fontenc}

\usepackage[utf8]{inputenc}

\usepackage{microtype}

\usepackage{inconsolata}

\usepackage{amsmath}
\usepackage{amssymb}
\usepackage{mathtools}
\usepackage{tabularx}
\usepackage{booktabs}
\usepackage{multirow}
\usepackage{bbding}
\usepackage{pifont}
\usepackage{wasysym}
\usepackage{amsfonts}
\usepackage{float}
\usepackage{mathtools, nccmath}
\usepackage{subcaption}

\usepackage{microtype}
\usepackage{graphicx}
\usepackage{caption}
\usepackage{subcaption}
\usepackage[inline,shortlabels]{enumitem}
\usepackage[ruled,vlined]{algorithm2e}
\usepackage{xurl} 

\graphicspath{{./figures}}

\usepackage{geometry}
\usepackage[permil]{overpic}

\usepackage{tikz}
\usetikzlibrary{shapes}

\usepackage{siunitx}  
\def \figurename {Fig.}

\def \figsnamelong {Figures}  

\def \Tablename {Table}
\def \Tabsname {Tables}
\def \appendixname {Appendix}

\newcommand{\figref}[1]{\figurename~\ref{#1}}

\newcommand{\Figsref}[2]{\figsnamelong~\ref{#1}--\ref{#2}}
\newcommand{\tabref}[1]{\Tablename~\ref{#1}}
\newcommand{\tabsref}[2]{\Tabsname~\ref{#1}--\ref{#2}}
\newcommand{\secref}[1]{§\ref{#1}} 
\newcommand{\appref}[1]{\appendixname~\ref{#1}}

\newcommand{\myeqref}[1]{Eq.~(\ref{#1})}  
\newcommand{\myeqsref}[2]{Eqs.~(\ref{#1})--(\ref{#2})}
\newcommand{\etc}{etc.\ }
\newcommand{\eg}{e.g., }

\newcommand{\ie}{i.e., }
\newcommand{\vs}{vs.\ }

\newcommand{\indexedset}[2]{\lbrace {#1} | {#2} \rbrace}

\newcommand{\nonfirstparagraph}[1]{\paragraph{#1}}

\usepackage{colortbl}
\usepackage{xcolor, soul}
\definecolor{myorange}{RGB}{251, 230, 207}
\definecolor{myblue}{RGB}{220, 232, 250}
\definecolor{mybluedark}{RGB}{79, 92, 156}
\definecolor{myreddark}{RGB}{192, 84, 95}
\definecolor{mypink}{RGB}{241, 207, 205}
\definecolor{mygreen}{RGB}{216, 231, 214}
\definecolor{mygreendark}{RGB}{95, 157, 104}

\newcommand{\chris}[1]{#1}

\newcommand{\model}{\texttt{IDAS}}
\newcommand{\neighbors}[2]{\mathcal{N}_{#1}(#2)}
\newcommand{\data}{\mathcal{D}_x}
\newcommand{\kmeans}{$K$-means}
\newcommand{\stdev}[2]{#1_{\pm#2}}

\DeclareMathOperator*{\argmax}{argmax}
\DeclareMathOperator*{\instruction}{\texttt{inst}}

\makeatletter
\def\Url@twoslashes{\mathchar`\/\@ifnextchar/{\kern-.2em}{}}
\g@addto@macro\UrlSpecials{\do\/{\Url@twoslashes}}
\makeatother

\usepackage{stfloats} 

\title{{\model}: Intent Discovery with Abstractive Summarization}

\author{{\centering 
    Maarten De Raedt$^{\diamondsuit\clubsuit}$~ 
    Fréderic Godin$^\diamondsuit$~
    Thomas Demeester$^\clubsuit$~
    Chris Develder$^\clubsuit$}  \\
    $^\diamondsuit$ Sinch Chatlayer ~~$^\clubsuit$ Ghent University \\
    \texttt{\{maarten.deraedt, thomas.demeester, chris.develder\}@ugent.be}  \\
    \texttt{frederic.godin@sinch.com} \\
}

\begin{document}
\maketitle
\begin{abstract}
Intent discovery is the task of inferring latent intents from a set of unlabeled utterances, and is a useful step towards the efficient creation of new conversational agents.
We show that recent competitive methods in intent discovery can be outperformed by clustering utterances based on \emph{abstractive summaries},
\ie``labels'', that retain the core elements while removing non-essential information.
We contribute the {\model} approach, which collects a set of descriptive utterance
labels by prompting a Large Language Model, starting from a well-chosen seed set of prototypical utterances, to bootstrap an In-Context Learning procedure to generate labels for non-prototypical utterances. The utterances and their resulting noisy labels are then encoded by a \emph{frozen} pre-trained encoder, and subsequently clustered to recover the latent intents.
For the \emph{unsupervised} task (without any intent labels) {\model} outperforms the state-of-the-art by up to +7.42\% in standard cluster metrics for the Banking, StackOverflow, and Transport datasets.
For the \emph{semi-supervised} task (with labels for a subset of intents) {\model} surpasses 2 recent methods on the CLINC benchmark without even using labeled data. 
\end{abstract}

\section{Introduction}
Intent classification is ubiquitous in conversational modelling. To that end, finetuning Large Language Models (LLMs) on task-specific intent data has been proven very effective \cite{casanueva-etal-2020-efficient, zhang-etal-2021-effectiveness-pre}.
However, such finetuning requires manually annotated (utterance, intent) pairs as training data, which 
are time-consuming and thus expensive to acquire.
Companies often have an abundance of utterances relevant to the application area of their interest, \eg those exchanged between customers and support agents, but manually annotating them remains costly. 
\begin{table}[t!]
\scriptsize
\centering
\addtolength{\tabcolsep}{-3pt}
\begin{tabular}{p{4.7cm} p{2.5cm}}
\toprule
Utterance &  Generated label \\
\midrule
find out when my next upcoming payday will be & when is next payday \\
my next paycheck is available when & when is next payday \\
what is the date of my last paycheck & when was last payday \\ \midrule
i want to know how to change my oil & how to change oil \\
what is the way to change motor oil & how to change oil \\ 
how easy is it to change your own oil & DIY oil change \\ \midrule
can you tell me the \emph{apr} on my visa card & \emph{interest rate inquiry} \\
what's the annual rate on my discover card & interest rate inquiry \\
\bottomrule
\end{tabular}
\addtolength{\tabcolsep}{3pt}
\caption{\emph{Illustration} based on GPT-3 and CLINC \cite{larson-etal-2019-evaluation}, demonstrating how \emph{abstractly} summarizing utterances retains the core elements while removing non-intent related information.
The example in the bottom block, where \emph{apr} is labeled as \emph{interest rate inquiry}, exemplifies the broad domain knowledge captured by LLMs.}
\label{table:placeholder}
\end{table}
Consequently, intent discovery aims to
recover latent intents without 
using any such manually annotated utterances, 
by partitioning a given set of (unlabeled) utterances into clusters, where utterances within a cluster should 
share the same \emph{conversational goal} or \emph{intent}.

Prior works typically
\begin{enumerate*}[(i)]
\item train an unsupervised sentence encoder to map utterances to vectors, after which these are
\item clustered to infer latent intents.
\end{enumerate*}
Such unsupervised encoder training is achieved largely under the assumption that 
utterances with similar encodings convey the same intent. For instance, by iteratively clustering and updating the encoder with supervision from the cluster assignments~\cite{xie2016unsupervised, caron2018deep, hadifar-etal-2019-self, Zhang_Xu_Lin_Lyu_2021}, or by retrieving utterances with similar encodings and using them as positive pairs to train the encoder with contrastive learning~\cite{zhang-etal-2021-supporting, zhang-etal-2022-new}.

Yet, it remains unclear which particular features cause utterance representations to be similar. Various noisy features unrelated to the underlying intents, \eg \emph{syntax}, \emph{n-gram overlap}, \emph{nouns}, \etc may contribute in making utterances similar, leading to sentence encoders whose vector encodings may inadequately represent the underlying intents.

Different from prior works that train unsupervised encoders, we use a pre-trained encoder without requiring any further finetuning, since we propose making utterances more (dis)similar in the textual space by \emph{abstractly} summarizing them into concise descriptions, \ie ``labels'', that preserve their core elements while removing non-essential information. We hypothesize that these core elements better represent intents and prevent non-intent related information from influencing the vector similarity. \tabref{table:placeholder} illustrates how labels retain the intent-related information by discarding irrelevant aspects such as syntax and nouns. 

This paper introduces Intent Discovery with Abstractive Summarization ({\model} in short), whereby the label generation process builds upon recent advancements of In-Context Learning (ICL)~\cite{brown2020language}. 
In ICL, an LLM is prompted with an instruction including a small number of (input, output) demonstrations of the task at hand. ICL has shown to be effective at few-shot learning without additional LLM finetuning \cite{min-etal-2022-noisy, min-etal-2022-metaicl}. However, intent discovery is unsupervised and therefore lacks the annotated (utterance, label) demonstrations required for ICL. To overcome this limitation, our proposed {\model} proceeds in four steps. First, a subset of diverse \emph{prototypical} utterances representative of distinct latent intents are identified by performing an initial clustering and selecting those utterances closest to each cluster's centroid, for which an LLM is then prompted to generate a short descriptive label. Second, labels for the remaining \emph{non-prototypical} utterances are obtained by retrieving the subset of the $n$ utterances most similar to the input utterance, from the continually expanding set of utterances with already generated labels (initialized with just the prototypes), and using those $n$ neighbors as ICL-demonstrations to generate the input utterance's label. Third, as the generated labels may still turn out too general or noisy, utterances with their labels are combined into a single vector representation using a \emph{frozen} pre-trained encoder.  
Finally, {\kmeans} clusters the combined encodings to infer latent intents.

We compare our {\model} approach with the state-of-the-art in unsupervised intent discovery on Banking \cite{casanueva-etal-2020-efficient}, StackOverflow \cite{xu-etal-2015-short}, and a private dataset from a  transport company, to assess
{\model}'s effectiveness in practice. We show that {\model} substantially outperforms the state-of-the-art, with average improvements in cluster metrics of +3.94\%, +2.86\%, and +3.34\% in Adjusted Rand Index, Normalized Mutual Information, and Cluster Accuracy, respectively.
Further, {\model} surpasses two \emph{semi-supervised} intent discovery methods on CLINC~\cite{larson-etal-2019-evaluation} despite 
not using any ground truth annotations. 

\section{Related Work}

\begin{figure*}
    \centering
    \includegraphics[width=\textwidth]{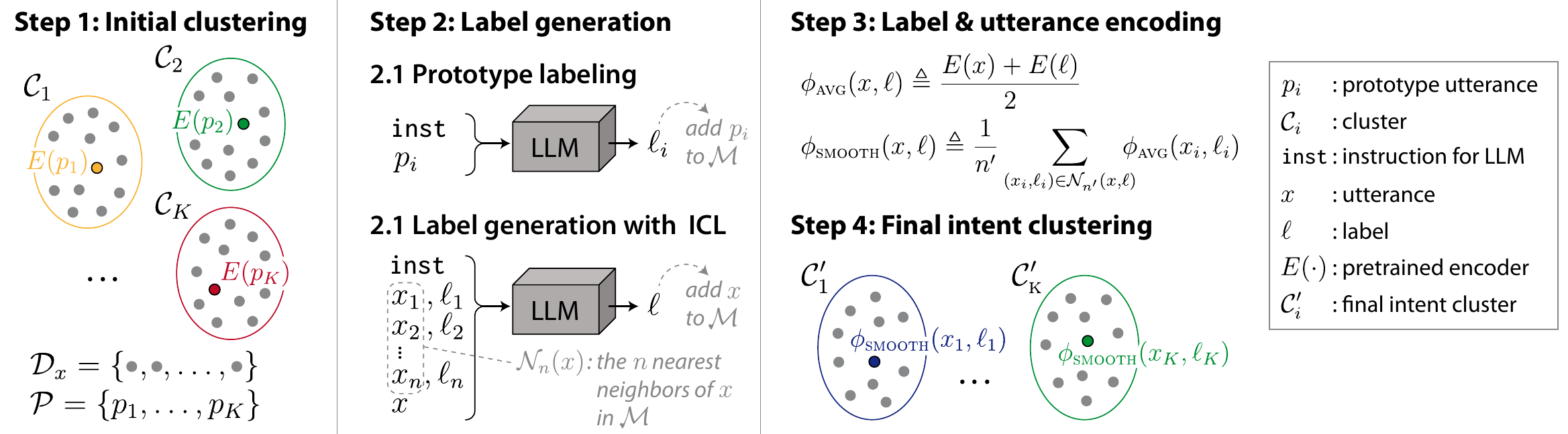}
    \caption{Overview of our {\model} approach.} 
    \label{fig:method-overview}
\end{figure*}

\paragraph{Statistical approaches:}
Early, more general short text clustering methods employ statistical methods such as tf-idf~\cite{sparck1972statistical}, to map text to vectors. Yet, the sparsity of these encodings prevents similar texts, but phrased with different synonyms, from being assigned to the same cluster. To specifically mitigate this synonym effect, external features have been used to enrich such \emph{sparse} vectors, \eg with WordNet~\cite{miller1995wordnet} synonyms or lexical chains~\cite{hotho2003ontologies,wei2015semantic}, or Wikipedia titles or categories \cite{banerjee2007clustering,hu2009exploiting}.

\nonfirstparagraph{Neural sentence encoders:}
Rather than relying on external knowledge sources, neural approaches pre-train sentence encoders in a self-supervised way 
\cite{kiros2015skip, gao-etal-2021-simcse}, or with supervision \cite{conneau-etal-2017-supervised, reimers-gurevych-2019-sentence,gao-etal-2021-simcse}, to produce \emph{dense} general-purpose vectors that better capture synonymy and semantic relatedness. 

\nonfirstparagraph{Unsupervised intent discovery:}
Since general-purpose neural encoders may fail to capture
domain-specific intent information, intent discovery solutions have shifted towards 
unsupervised sentence encoders specifically trained on the domain data at hand.
For instance, \citet{xu-etal-2015-short} train a self-supervised Convolutional Neural Network, and use it to encode and cluster utterances with {\kmeans}. \citet{zhang-etal-2022-new} adopt the same 2-step approach, but instead pre-train the encoder with contrastive learning, where utterances with similar vector encodings are retrieved to serve as positive pairs. A more common strategy is to cluster and train the encoder end-to-end, either by 
\begin{enumerate*}[(i)]
    \item \emph{iteratively} clustering utterances and updating the encoder with supervision from the cluster assignments \cite{xie2016unsupervised, caron-etal, hadifar-etal-2019-self}, or
    \item \emph{simultaneously} clustering utterances and updating the encoder's weights with a joint loss criterion \cite{yang2017towards, zhang-etal-2021-supporting}.
\end{enumerate*}

As an alternative strategy to make utterances more (dis)similar based on the intents they convey, we employ an LLM to summarize utterances into labels that retain both the utterances' core elements and domain-specific information as encoded in the LLM's weights.
Since our generated labels should increase the (dis)similarity of (un)related utterances in the input space, rather than directly in the vector space, we use a \emph{frozen} pre-trained encoder, thus deviating from the above methods that \emph{train} unsupervised encoders.

\nonfirstparagraph{Semi-supervised intent discovery:}
Similar to our current work, the aforementioned methods focus on \emph{unsupervised} intent discovery.
In the related but different \emph{semi-supervised} intent discovery task, 
a fraction of the latent intents is assumed to be known, 
\ie the ``Known Class Ratio''.
Annotated data from these known intents is exploited to improve the detection of both known and unknown intent utterances, \eg 
by optimizing a cluster loss 
with pairwise constraints derived from utterances of the same known intent \cite{lin2020discovering}.
Alternative 
2-step approaches first pre-train encoders with supervision from known intent utterances, then either directly encode and cluster utterances with {\kmeans} \cite{shen-etal-2021-semi}, or further refine the encoder on the unlabeled utterances. 
The latter refinement can be achieved through contrastive learning \cite{zhang-etal-2022-new} or by iteratively clustering and updating the encoder \cite{zhang-etal-2021-textoir, Zhang_Xu_Lin_Lyu_2021}.

\nonfirstparagraph{In-context learning:} 
The core idea of ICL \cite{brown2020language} is to perform tasks 
through inference, \ie without updating parameters,
by prompting an LLM with the string concatenation comprising
\begin{enumerate*}[(i)]
    \item a task instruction,
    \item a small set of (input, output) demonstrations, and
    \item the input.
\end{enumerate*} 
We implement {\model}'s label generation process with ICL, as it has shown to substantially outperform zero-shot approaches \emph{without} demonstrations \cite{min-etal-2022-noisy, min-etal-2022-metaicl, chen-etal-2022-meta}.
However, since we focus on unsupervised intent discovery and thus lack annotated (utterance, label) demonstrations, we bootstrap the set of demonstrations with automatically retrieved ``prototypes''.
Rather than selecting demonstrations randomly, \citet{liu-etal-2022-makes} found that it is more effective to 
pick demonstrations similar to the input utterance, which we thus 
do.
Note that alternative methods are possible \cite{rubin-etal-2022-learning, sorensen-etal-2022-information}.

\section{Methodology}
\label{section:methodology}
\paragraph{Task formulation:} Let $\indexedset{(x_i,y_i)}{i=1\ldots N}$ be a dataset of $N$
utterances $x \in \mathcal{X}$ from the set of natural language expressions $\mathcal{X}$,
with corresponding intents $y$  chosen from a set of $K$ possible intents
 $\mathcal{Y} = \indexedset{y_i}{i=1\ldots K}$. 
Given the utterances without the intents,
$\data{} = \indexedset{x_i}{i=1\ldots N}$, 
intent discovery aims to infer $\mathcal{Y}$ from $\data{}$ by mapping utterances $x_i$ 
to vectors $E(x_i)$ with encoder $E: \mathcal{X} \to \mathbb{R}^d$, based on which the utterances are partitioned into clusters $\indexedset{\mathcal{C}_i}{i=1 \ldots K}$, such that clustered utterances (\eg $x_{i,j}, x_{k,j} \in \mathcal{C}_j$) share the same intent ($y_{i,j} = y_{k,j}$), while utterances from different clusters (\eg $x_{i,j} \in \mathcal{C}_j$ and $x_{k,l} \in \mathcal{C}_l$, $\mathcal{C}_l \neq \mathcal{C}_k$)
have distinct intents ($y_{i,j} \neq y_{k,l}$).

\paragraph{Overview:}
As summarized in \figref{fig:method-overview}, to 
infer latent intents {\model}
\begin{enumerate*}[(1)]
    \item identifies a subset of diverse ``prototypes'', $\mathcal{P} \subset$ $\data{}$, representative of the latent intents (\secref{subsec:step1}); then
    \item independently summarizes them into labels, which are further used to also generate 
    labels for the remaining non-prototypical utterances $x\in\data{}\setminus\mathcal{P}$, by retrieving from the subset $\mathcal{M}$ of utterances that already have labels (initially $\mathcal{P}$) the set $\neighbors{n}{x}$ of $n$ utterances most similar to $x$ as ICL-demonstrations for generating the label of $x$ (\secref{subsec:step2}); further
    \item encodes utterances and their labels into a single vector representation with a \emph{frozen} pre-trained encoder (\secref{subsec:step3}); and finally
    \item infers the latent intents by performing {\kmeans} on the combined representations (\secref{subsec:step4}).
\end{enumerate*}

\subsection{Step~1: Initial Clustering}
\label{subsec:step1}

The objective of this step is to identify a diverse set of prototypes, $\mathcal{P} \subset \data{}$, that in Step~2 will be automatically labeled by an LLM and serve as initial demonstrations for generating the labels of non-prototypical utterances.
It is therefore important to choose
prototypes $p \in \mathcal{P}$ that each represent a distinct latent intent $y \in \mathcal{Y}$, and collectively cover as many as possible of all latent intents.
We assume a similarity function between two vector representations of utterances by $s: \mathbb{R}^d \times \mathbb{R}^d \to \mathbb{R}$, and use it to retrieve prototypes by performing an initial clustering on the utterances in ${\data}$, in the vector representation space induced by encoder $E$. Then we select a prototype from each identified cluster, as the utterance in that cluster whose vector representation is closest to the cluster's centroid.

Formally, the utterances in ${\data}$ are first encoded with $E$ and then partitioned into $K$ (=$|\mathcal{Y}|$) clusters
\begin{equation*}
 \mathcal{C}_1, \ldots, \mathcal{C}_K = K\textnormal{-means}(\data{}),
\end{equation*} for which the respective centroids $c_i \in \mathbb{R}^d$ and prototypes $p_i \in \data$ are calculated as
\begin{equation*}
 c_i = \frac{1}{|\mathcal{C}_i|}\sum_{x \in \mathcal{C}_i} E(x), \quad 
 p_i = \argmax_{x \in \mathcal{C}_i} s(E(x), c_i).
\end{equation*}
\subsection{Step~2: Label Generation}\label{subsec:step2}
\paragraph{Step~2.1: Prototype Labeling}
To generate label $\ell_i$ for prototype $p_i$, we employ an LLM and provide it with an instruction ($\instruction$) such as ``\texttt{describe the question in a maximum of 5 words}''.
The LLM then generates a concise description of the prototype $p_i$, which we use as its label $\ell_i$.
Mathematically, this is represented as
\begin{equation*}
\ell_i = \argmax_{\ell \in \mathcal{X}}P(\ell|\instruction, p_i),
\end{equation*}
\noindent where $P$ denotes the probability distribution of the LLM, and $\ell_i$ represents the token sequence $t_{1_i}, \ldots, t_{l_i}$ output by the LLM.

\nonfirstparagraph{Step~2.2: Label Generation with ICL}
To generate label $\ell$ for the non-prototypical utterance $x \in \data{}\setminus\mathcal{P}$, {\model} utilizes ICL by conditioning an LLM on the prompt, \ie the string concatenation of 
\begin{enumerate*}[(i)]
    \item an instruction $\instruction$, \eg ``\texttt{classify the question into one of the labels}'',
    \item the set of $n$ demonstrations of (utterance, label) pairs
    $\indexedset{(x_i, \ell_i)}{i=1\ldots n}$, and
    \item the utterance $x$ itself.
\end{enumerate*}
Formally, the label is the token sequence generated by the LLM that maximizes the probability given the prompt:
\begin{equation*}
    \ell = \argmax_{\ell \in \mathcal{X}}\,P(\ell|\instruction,x_1,\ell_1, \ldots, x_n, \ell_n, x).
\end{equation*}
Since unsupervised intent discovery lacks manually annotated demonstrations, {\model} uses a continually expanding set of utterances with \emph{automatically} generated labels, denoted by $\mathcal{M}$.
Initially, $\mathcal{M} = \mathcal{P}$, with $\mathcal{P}$ the set of prototypes from
Step~2.1. An utterance $x$ with newly generated label $\ell$ is added to $\mathcal{M}$, such that it can serve as a demonstration for remaining unlabeled utterances.

Typically, ICL uses a small set of $n$ demonstrations
\begin{enumerate*}[(i)]
\item due to the limit on the number of input tokens of LLMs, and 
\item because performance does not improve for larger number of demonstrations \cite{min-etal-2022-rethinking}.
\end{enumerate*}
Moreover, \citet{liu-etal-2022-makes} found that selecting demonstrations as samples similar to the test input, rather than choosing them randomly, substantially boosts ICL's performance.
Therefore, {\model} adopts \textsc{kate} \cite{liu-etal-2022-makes} by first mapping utterances in $\mathcal{M}$ to vectors with encoder $E$, and then using the similarity function $s$ to select the set of the $n$ most similar utterances\footnote{We set hyperparameter $n$ to 8, based on the findings of \citet{min-etal-2022-rethinking, lyu2022z}. Ablations for different $n$ values are presented in \secref{subsec:ablations}.} from $\mathcal{M}$ to $E(x)$, denoted by $\neighbors{n}{x} \subset \mathcal{M}$, as demonstrations for input utterance $x$.

Note that while we use ``\texttt{classify}'' in the instruction, we do not consider the prototypical labels generated in Step~1 as a fixed label set (\ie \emph{verbalizers}).
Rather, label $\ell$ for non-prototypical utterance $x$ is the token sequence as generated directly by the LLM.
As a result, labels for non-prototypical utterances may still differ from those generated for the prototypes.
Particularly, 
the LLM can generate new labels for input utterances that represent intents for which no prototypes have been identified yet, and thus have no ICL demonstrations of the latent intent.
Thus, we minimize error propagation from Step~1.
On the other hand, when the LLM considers that a demonstration likely shares the same latent intent with the input utterance, the ``\texttt{classify}'' instruction should encourage the LLM to generate a copy of that demonstration's label, which in turn minimizes variation among generated labels of utterances with the same latent intent.

\subsection{Step 3: Encoding Utterances and Labels}
\label{subsec:step3}
After Step~2, each utterance $x \in {\data}$  
has an associated generated label
$\ell \in \mathcal{M}$.
We use the pre-trained encoder $E$ to respectively encode the utterances and their corresponding labels into separate vectors $E(x)$ and $E(\ell)$, after which these are averaged into the combined representation:  
\begin{equation}\label{equation_encoding}
    \phi_{\textsc{avg}}(x, \ell) \triangleq 
    \frac{E(x) + E(\ell)}{2}.
\end{equation}
(Note that utterances could also be represented just by their label encoding $E(\ell)$, yet such generated labels
could be noisy or overly general.)  

We further contribute a non-parametric smoothing method that  
\begin{enumerate*}[(i)]
    \item aims to suppress features that are specific to individual utterances and thus potentially less representative of the underlying intents, while
    \item enhancing those features that are shared across utterances and thus more likely to be representative of the latent intents.
\end{enumerate*} 
We therefore represent utterance $x$ as the average of the vector encodings  
of the $n'$ most similar 
utterances $\neighbors{n'}{x,\ell}$ to $x$, including $x$ itself:

\begin{equation}\label{equation_avg_encoding}
\phi_{\textsc{smooth}}(x,\ell) \triangleq \frac{1}{n'} \sum_{(x_i,\ell_i) \in \neighbors{n'}{x,\ell}} \phi_{\textsc{avg}}(x_i, \ell_i).
\end{equation}
We automatically determine the value of $n'$ 
as the value that maximizes the average silhouette score \cite{rousseeuw1987silhouettes} among all samples, which for sample $i$ is given by 
\begin{equation*}\label{equation_silhouette}
\textrm{silhouette-score}(i) = \frac{b(i)-a(i)}{\max(a(i), b(i))}, 
\end{equation*}
where $a(i)$ is the average distance of sample $i$ to all other samples in its cluster, and $b(i)$ is the average distance of sample $i$ to all samples in the neighboring cluster nearest to $i$.

\subsection{Step 4: Final intent discovery}
\label{subsec:step4}
To finally infer the latent intents, we represent each utterance $x \in {\data}$ with its  label $\ell$ as $\phi_{\textsc{smooth}}(x,\ell)$, 
and apply {\kmeans} clustering, setting $K$ to the ground truth number of latent intents $|\mathcal{Y}|$, following \citet{hadifar-etal-2019-self, zhang-etal-2021-supporting, Zhang_Xu_Lin_Lyu_2021, zhang-etal-2022-new}.

\section{Experimental Setup}
\label{sec:experiment-setup}

\subsection{Datasets}
We evaluate our {\model} approach on two widely adopted intent classification datasets, CLINC \cite{larson-etal-2019-evaluation} and Banking \cite{casanueva-etal-2020-efficient}, as well as the StackOverflow topic classification dataset \cite{xu-etal-2015-short}.
We also use a private dataset from a transportation company. \tabref{table:datasets} summarizes dataset statistics.

\begin{table}
\smaller
\centering
\addtolength{\tabcolsep}{4pt}
\begin{tabular}{lccc}
\toprule
Dataset & \#\,Train & \#\,Test & \#\,Intents \\
\midrule
CLINC & 18,000 & 2,250 & 150 \\
Banking & 9,016 & 3,080 & 77 \\
Transport & - & 1,257  & 42 \\
StackOverflow & 18,000 & 1,000 & 20 \\
\bottomrule
\end{tabular}
\addtolength{\tabcolsep}{-4pt}
\caption{
Dataset statistics.}
\label{table:datasets}
\end{table}

\subsection{Baselines}
On Banking, StackOverflow, and our Transport dataset, we compare {\model} against the state-of-the-art in unsupervised intent discovery,
\ie the MTP-CLNN model \cite{zhang-etal-2022-new}  that outperforms prior unsupervised methods, such as DEC \cite{pmlr-v48-xieb16}, DCN \cite{pmlr-v70-yang17b}, and DeepCluster \cite{caron-etal}.
As the MTP-CLNN model is pre-trained on the annotated training data of CLINC, directly comparing against it would be unfair. Instead, we compare our approach on CLINC with two state-of-the-art \emph{semi-supervised} intent discovery methods, DAC \cite{Zhang_Xu_Lin_Lyu_2021} and SCL+PLT \cite{shen-etal-2021-semi}. 
Compared to the semi-supervised setting, the unsupervised setting without annotations is thus 
more challenging. We report results of DAC and SCL+PLT with an increasing ``Known Class Ratio'' (KCR) of 25\%, 50\%, and 75\%, using the annotated data for the known intents of \citet{shen-etal-2021-semi}.

\subsection{Evaluation} Following \citet{Zhang_Xu_Lin_Lyu_2021, shen-etal-2021-semi, zhang-etal-2022-new}, we assess cluster performance by comparing the predicted clusters to the ground truth intents using the
\begin{enumerate*}[(i)]
    \item Adjusted Rand Index (ARI) \cite{steinley2004properties},
    \item Normalized Mutual Information (NMI), and
    \item Cluster Accuracy (ACC) based on the Hungarian algorithm \cite{kuhn1955hungarian}.
\end{enumerate*} 
Since {\model}'s label generation process may depend on the order in which utterances occur, we perform
Steps~1--2 leading to utterance labels 5 times, shuffling the utterance order.
We further conduct the final clustering Step~4 with 10 different seeds for each of those
5 label generation runs, to account for variation incurred by {\kmeans}.
For each dataset, we then average the results in terms of means and standard variations across each of these 5 sets.

\subsection{Implementation}
\paragraph{Encoder:} We use the same pre-trained encoder $E$ in all steps of our approach, \ie to 
\begin{enumerate*}[(i)]
    \item retrieve prototypes (\secref{subsec:step1}),
    \item mine the $n$ demonstrations $\neighbors{n}{x}$ for utterance $x$ (\secref{subsec:step2}), and
    \item encode utterances with their labels using \myeqsref{equation_encoding}{equation_avg_encoding} (\secref{subsec:step3}).
\end{enumerate*}
To rule out performance differences stemming purely from the encoder,
we employ the same pre-trained encoder as the baseline we compare with: we use
the MTP encoder for Banking, StackOverflow, and Transport, where we compare to MTP-CLNN \cite{zhang-etal-2022-new}, and the SBERT encoder \texttt{paraphrase-mpnet-base2} (\ie SMPNET) \cite{reimers-gurevych-2019-sentence} for CLINC, where we compare to DAC \cite{Zhang_Xu_Lin_Lyu_2021} and SCL+PLT \cite{shen-etal-2021-semi}.

\nonfirstparagraph{Language models and prompts:} {\model} uses the \texttt{text-davinci-003} GPT-3 model \cite{ouyangtraining} as its LLM for label generation.
We adopt the OpenAI playground default values, except for the temperature, which we set to $0$ to minimize variation among generated labels of utterances with the same latent intent.
To generate prototypical labels (\secref{subsec:step2}), we use the instruction ``\texttt{Describe the {domain} question in a maximum of 5 words}'', where the domain is \emph{banking}, \emph{chatbot}, or \emph{transport} for the corresponding dataset.
Since StackOverflow is a topic rather than an intent classification dataset, we adopt a slightly different prototypical prompt. 
To generate labels for non-prototypical utterances with ICL (\secref{subsec:step2}), we use 
``\texttt{Classify the {domain} question into one of the provided labels}'' for all 4 datasets.
See \appref{appendix:prompts} for full prompts and examples.

\nonfirstparagraph{Nearest neighbor retrieval:} 
The function $s$ is implemented with cosine similarity.
We use $n$\,=\,8 demonstrations $\neighbors{n}{x}$ to generate label $\ell$ for utterance $x$ (\secref{subsec:step2}), based on \citet{min-etal-2022-rethinking} and \mbox{\citet{lyu2022z}}, who report that further increasing $n$ does not improve ICL's performance. The number of smoothing samples $n'$ 
is determined by running the final {\kmeans} (\secref{subsec:step4}) multiple times with 
$n'$ ranging from 5 to 45 and selecting 
the value that maximizes the average silhouette score. 

\section{Results and Discussion}
\label{sec:experiments}
\begin{table*}[t]
\centering
\smaller
\setlength\tabcolsep{2.4pt} 
\begin{tabular}{l lll lll lll lll}
\toprule
& \multicolumn{3}{c}{Banking}  & \multicolumn{3}{c}{StackOverflow} & \multicolumn{3}{c}{Transport} & \multicolumn{3}{c}{Average} \\ 
\cmidrule(lr){2-4} \cmidrule(lr){5-7} \cmidrule(lr){8-10} \cmidrule(lr){11-13}
Model & ARI & NMI & ACC  & ARI & NMI & ACC & ARI & NMI & ACC & ARI & NMI & ACC \\
\midrule
MTP\textsuperscript{$\diamondsuit$} & 47.33 & 77.32 & 57.99 & 48.71 & 63.85 & 66.18 & - & - & - & 48.02 & 70.59 & 62.09 \\ 
MTP-CLNN\textsuperscript{$\diamondsuit$} & \underline{55.75} & \underline{81.80} & \underline{65.90} & \underline{67.63} & \underline{78.71} & \underline{81.43}  & - & - & - & \underline{61.69} & \underline{80.26} & \underline{73.67} \\ 
{\model} & \textbf{57.56} & \textbf{82.84} & \textbf{67.43} & \textbf{72.20} & \textbf{81.26} & \textbf{83.82}  & - & - & -  & $\stdev{\textbf{64.88}}{1.07}$ & $\stdev{\textbf{82.05}}{0.68}$ & $\stdev{\textbf{75.63}}{0.82}$ \\
\midrule
MTP\textsuperscript{$\spadesuit$} & 39.52 & 72.03 & 51.66 & 29.66 & 47.46 & 48.97  & 44.51 & 74.71 & 57.51 & $\stdev{37.90}{0.48}$ & $\stdev{64.73}{0.31}$ & $\stdev{52.69}{0.60}$ \\
MTP-CLNN\textsuperscript{$\spadesuit$} & \underline{52.47} & \underline{79.46} & \textbf{64.06} & \underline{62.53} & \underline{73.52} & \underline{78.82} & \underline{50.33} & \underline{77.77} & \underline{61.60}  & $\stdev{\underline{55.11}}{1.32}$ & $\stdev{\underline{76.92}}{0.74}$ & $\stdev{\underline{68.16}}{1.02}$ \\ 
{\model} & \textbf{53.31} & \textbf{80.43} & \underline{63.77} & \textbf{66.08} & \textbf{77.25} & \textbf{82.11} & \textbf{57.75} & \textbf{81.66} & \textbf{68.51} & $\stdev{\textbf{59.05}}{1.92}$ & $\stdev{\textbf{79.78}}{0.91}$ & $\stdev{\textbf{71.46}}{1.57}$ \\ 
\midrule
$\Delta$MTP-CLNN$^\diamondsuit$ & $+$1.81 & $+$1.04 & $+$1.53 & $+$4.57 & $+$2.55 & $+$2.39  & - & - & - & $+$3.19 & $+$1.79 & $+$1.96 \\
$\Delta$MTP-CLNN$^\spadesuit$ & $+$0.84 & $+$0.97 & $-$0.29 & $+$3.55 & $+$3.73 & $+$3.39  & $+$7.42 & $+$3.89 & $+$6.91 & $+$3.94 & $+$2.86 & $+$3.34 \\ \bottomrule
\end{tabular}
\caption{Comparison against \emph{unsupervised} state-of-the-art. $\diamondsuit$: results from \citet{zhang-etal-2022-new}. $\spadesuit$: results from (pre-)training MTP(-CLNN) on the test set (rather than a distinct unlabeled training set). The \textbf{best} model is typeset in bold and the \underline{runner-up} is underlined. $\Delta$MTP-CLNN values are the absolute gains of our {\model}.}\label{table:main_unsupervised}
\end{table*}

\begin{table}
\centering
\smaller
\setlength\tabcolsep{3.5pt} 
\begin{tabular}{ll lll} 
\toprule
& & \multicolumn{3}{c}{CLINC} \\ 
\cmidrule(lr){3-5}
KCR & Model & ARI & NMI & ACC \\
\midrule
0\% & SMPNET & 63.82 & 89.01 & 71.30 \\
& {\model} & 79.02$_{\pm1.14}$ & 93.82$_{\pm0.38}$ & 85.48$_{\pm0.84}$ \\
\midrule
25\% & DAC$^{\heartsuit}$ & 65.36 & 89.12 & 75.20 \\
& SCL+PLT$^{\heartsuit}$ & 64.78 & 89.31 & 73.77 \\
\midrule
50\% & DAC$^{\heartsuit}$ & 72.26 & 91.50 & 80.70 \\
& SCL+PLT$^{\heartsuit}$ & 73.25 & 92.21 & 80.59 \\
\midrule
75\% & DAC$^{\heartsuit}$ & 79.56 & 93.92 & 86.40 \\ 
& SCL+PLT$^{\heartsuit}$ & \textbf{83.44} & \textbf{95.25} & \textbf{88.68} \\
\bottomrule
\end{tabular}
\caption{Comparison against \emph{semi-supervised methods} DAC and SCL+PLT. $\heartsuit$: results from \citet{shen-etal-2021-semi}. Bold indicates \textbf{best} model. KCR: known class ratio.}
\label{table:main_semisupervised}
\end{table}

\subsection{Main Results}\label{subsec:main_results}
In \emph{unsupervised} clustering, no labels are available and thus there is only a test set, used to evaluate the model's induced clusters against gold standard labels \cite{xie2016unsupervised, yang2017towards, hadifar-etal-2019-self, zhang-etal-2021-supporting}.
In the \emph{semi-supervised} intent detection setting, intent labels are available for a subset of intents: there is an additional labeled training set --- which can be exploited, \eg for (pre-)training a sentence encoder.

\citet{zhang-etal-2022-new} evaluated their MTP and MTP-CLNN models by (pre-)training the encoder based on an unlabeled training set different from the test set
where (new) intent clusters are induced, \ie they evaluate on a held-out test set unseen during any (pre-)training phase.
Since in our {\model}, no encoder is trained, we perform Steps 1--4 on the (unlabeled) test set following \cite{xie2016unsupervised, yang2017towards, hadifar-etal-2019-self, zhang-etal-2021-supporting}. To ensure a fair comparison we also consider an MTP-CLNN that uses that same test set in (pre-)training its encoder (\ie for the $\mathcal{D}^\textrm{unlabeled}$ as defined in \citet{zhang-etal-2022-new}; results marked by $\spadesuit$ in \tabref{table:main_unsupervised}).
Note that the test sets for a particular dataset are identical across all reported results.

First, we compare {\model} against the state-of-the-art in the \emph{unsupervised} setting, \ie MTP-CLNN, with results reported in \tabref{table:main_unsupervised}.
Both in the original settings of \citet{zhang-etal-2022-new} (keeping the test data unseen during training, $\diamondsuit$) as well as when using the unlabeled test data in training MTP(-CLNN) ($\spadesuit$), our {\model} significantly surpasses it, with gains averaged over three datasets of +3.19--3.94\%, +1.79--2.86\% and +1.96--3.34\% in respectively ARI, NMI and ACC.
 We further find that {\model} consistently outperforms MTP-CLNN on all metrics and datasets,
 except for Banking, where {\model} and MTP-CLNN perform similarly (when comparing them in
 similar settings, \ie both using unlabeled test data in training phase).
Note that both {\model} and MTP-CLNN perform worse on StackOverflow and Banking
in our settings ($\spadesuit$) compared to the original results of \citet{zhang-etal-2022-new} ($\diamondsuit$), likely because in case of $\spadesuit$, the MPT(-CLNN) encoder(s) were trained 
on a substantially lower number of samples, \ie only 5.5\% for StackOverflow (1,000 for $\spadesuit$ \vs 18,000 for $\diamondsuit$) and 34\% for Banking (3,080 for $\spadesuit$ \vs 9,016 for $\diamondsuit$).

Second, we assess our {\model}'s performance in the \emph{semi-supervised} task setting, where a subset of intents has labeled data. Note however that our {\model} does not use the labels for those utterances in any way.
The results for CLINC presented in \tabref{table:main_semisupervised} show that {\model} outperforms both semi-supervised SCL+PLT and DAC methods for KCR's of 25\% and 50\%. Notably, {\model} surpasses SCL+PLT and DAC for KCR of 50\%, with improvements in the range of 
5.77--6.76\%, 1.61--2.32\%, and 4.78--4.89\% in ARI, NMI, and ACC, respectively.
Even for KCR\,=\,75\%, it performs just slightly worse than DAC, further
confirming {\model}'s effectiveness. 

\subsection{Ablations}
\label{subsec:ablations}
Below, we investigate the impact of
\begin{enumerate*}[(i)]
    \item\label{item:ablation2} the encoding strategies from \secref{subsec:step3}, and 
    \item\label{item:ablation1} ICL from \secref{subsec:step2}
\end{enumerate*} on {\model}'s performance.
The results for each ablation are averaged over 5 runs with the utterances' order corresponding to those used for presenting the main results, \ie with {\model}'s default parameters values.
Due to computation budget constraints, we only provide ablations on StackOverflow for~\ref{item:ablation1}, since it requires GPT-3.
For~\ref{item:ablation2}, we report results for Banking, StackOverflow, Transport, and CLINC.

\nonfirstparagraph{Effect of the encoding strategies:}
\begin{table*}
\centering
\smaller
\begin{tabular}{l ccc ccc ccc ccc}
\toprule
& \multicolumn{3}{c}{Banking}  & \multicolumn{3}{c}{StackOverflow} & \multicolumn{3}{c}{Transport} & \multicolumn{3}{c}{CLINC} \\ 
\cmidrule(lr){2-4} \cmidrule(lr){5-7} \cmidrule(lr){8-10} \cmidrule(lr){11-13}
Encoding & ARI & NMI & ACC  & ARI & NMI & ACC & ARI & NMI & ACC & ARI & NMI & ACC \\
\midrule 
$E(x)$ & 47.33 & 77.32 & 57.99 & 48.71 & 63.85 & 66.18  & 44.51 & 74.71 & 57.51 & 63.82 & 89.01 & 71.30\\
$E(\ell)$ & 52.45 & 81.14 & 62.31 & 67.94 & 80.60 & 80.05  & 54.37 & 80.68 & 64.66 & 75.01 & 93.04 & 81.27 \\ 
$\phi_{\textsc{avg}}(x,\ell)$ & 54.47 & 82.35 & 63.25 & 69.20 & 80.76 & 81.29  & 55.91 & 81.11 & 65.94 & 75.65 & 93.33 & 81.04 \\
$\phi_{\textsc{smooth}}(x,\ell)$ & \textbf{57.56} & \textbf{82.84} & \textbf{67.43} & \textbf{72.20}  & \textbf{81.26} & \textbf{83.82}  & \textbf{57.75} & \textbf{81.66} & \textbf{68.51} & \textbf{79.02} & \textbf{93.82} & \textbf{85.48} \\ 
\bottomrule
\end{tabular}
\caption{\emph{Effect of the encoding strategies.}}
\label{table:ablations_encoding}
\end{table*}

\begin{figure*}
    \centering
    \begin{subfigure}{0.268\textwidth}
      \includegraphics[width=\textwidth]{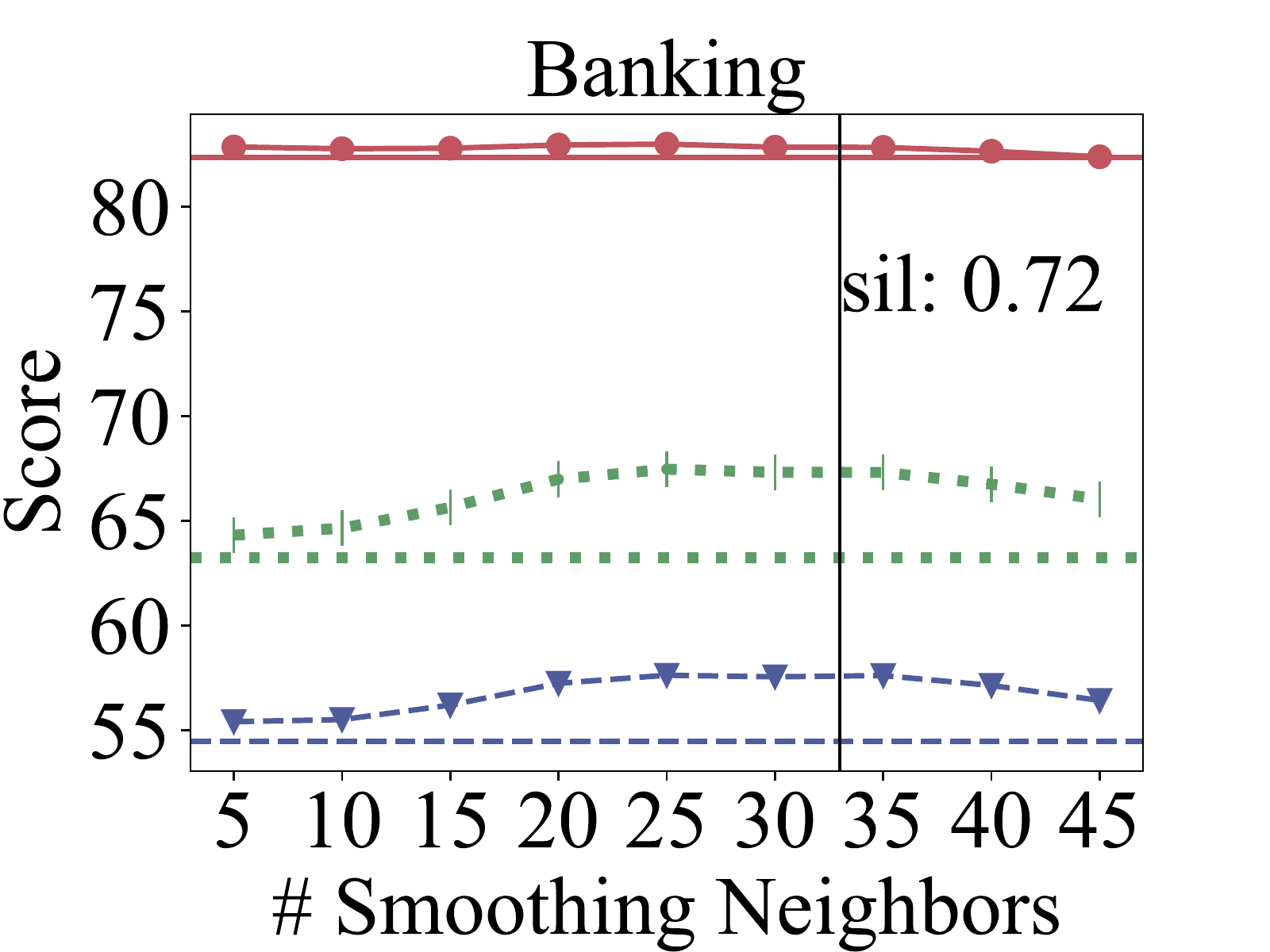}
    \end{subfigure}\hspace{-1.32em} 
    \begin{subfigure}{0.268\textwidth}
      \includegraphics[width=\textwidth]{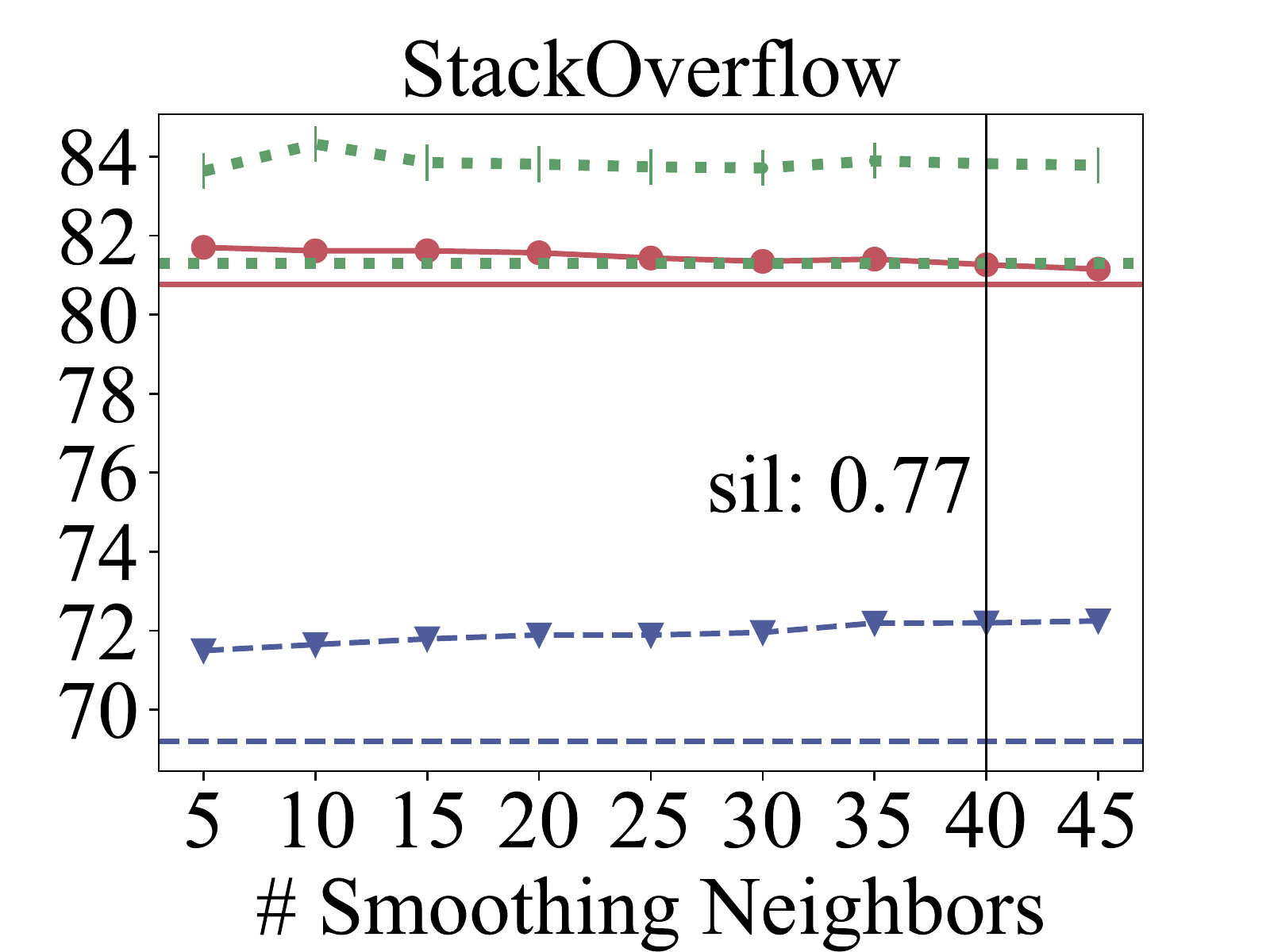}
    \end{subfigure}\hspace{-1.32em}
    \begin{subfigure}{0.268\textwidth}
      \includegraphics[width=\textwidth]{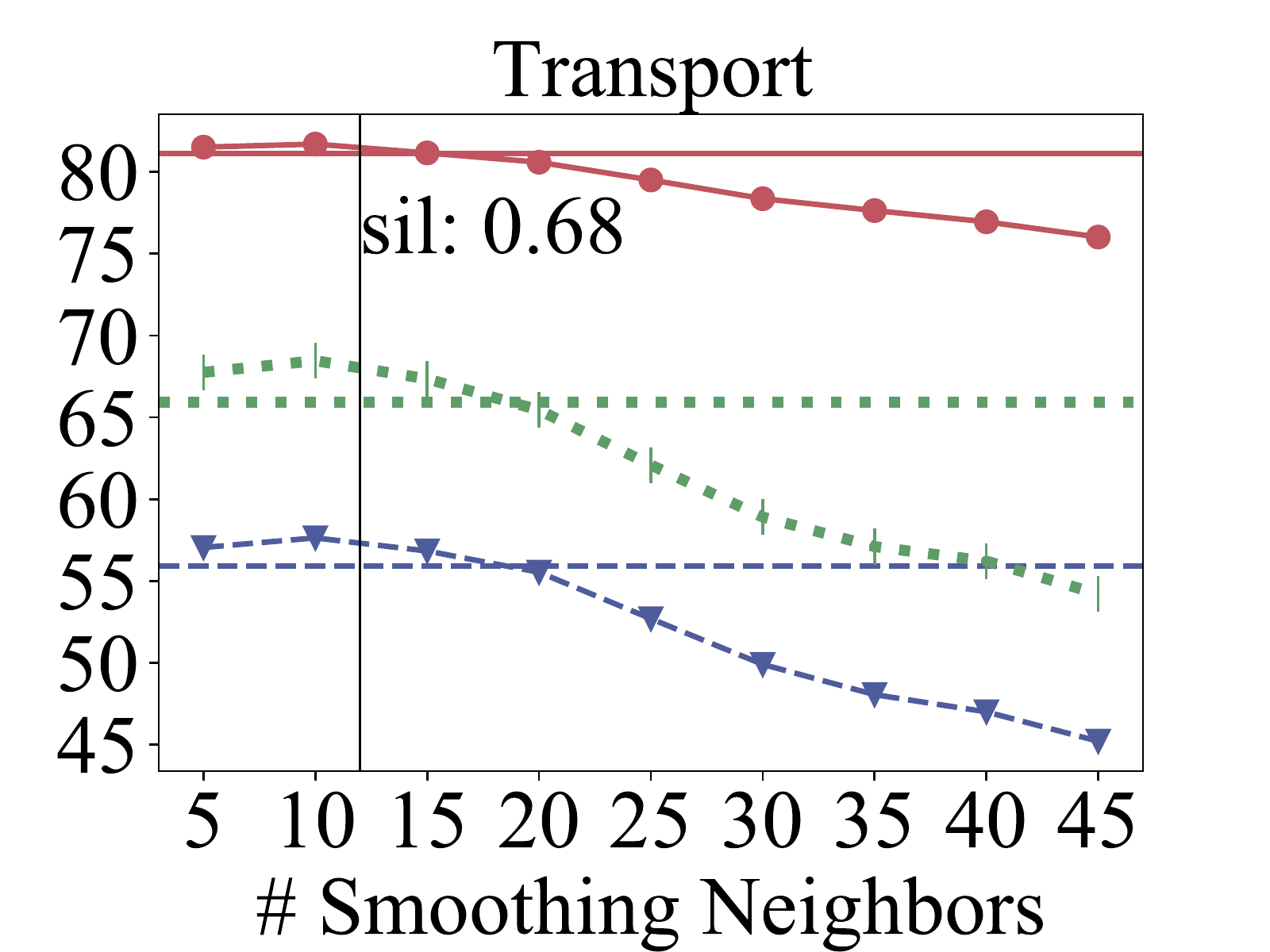}
    \end{subfigure}\hspace{-1.32em} 
    \begin{subfigure}{0.268\textwidth}
      \includegraphics[width=\textwidth]{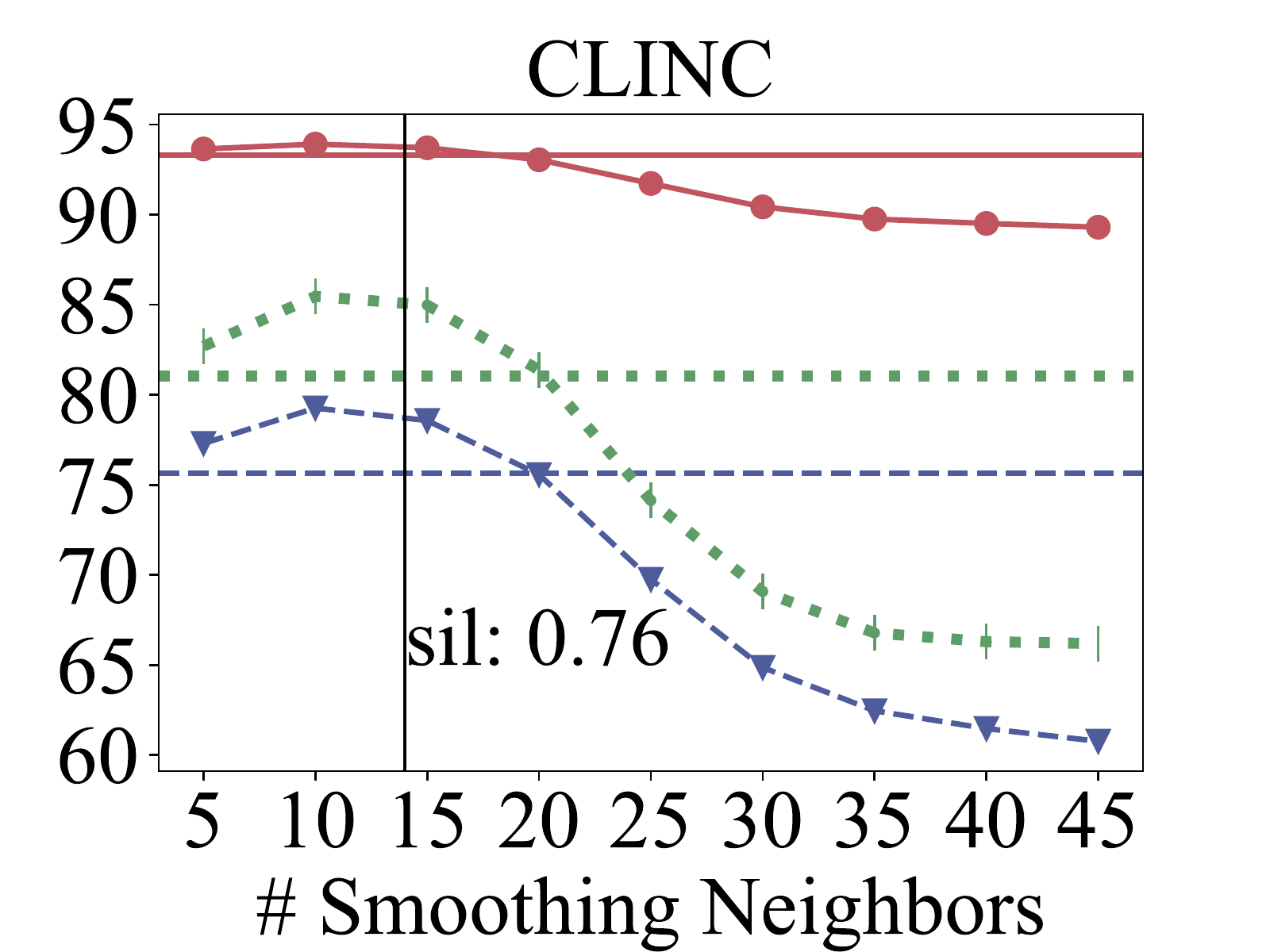}
    \end{subfigure}
    \begin{subfigure}{1.\textwidth}
      \centering
      \includegraphics[width=\textwidth]{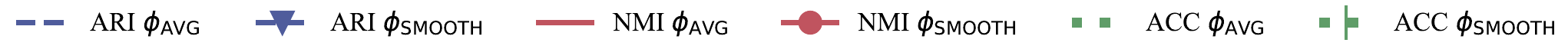}
      \label{fig:legend}
    \end{subfigure}\vspace{-1em}
    \caption{\emph{Inferring the number of smoothing neighbors $n'$}. The vertical lines represent the automatically determined number of smoothing neighbors corresponding to the highest silhouette score (sil).}
    \label{fig:smoothing_neighbors}
\end{figure*}

\tabref{table:ablations_encoding} compares 
the cluster performance of these four encoding strategies:
\begin{enumerate*}[(1)]
\item $E(x)$ encodes only utterances; 
\item $E(\ell)$ encodes only generated labels;
\item $\phi_{\textsc{avg}}(x,\ell)$ (\myeqref{equation_encoding}) averages utterance and label encodings into a single vector representation;
\item $\phi_{\textsc{smooth}}(x,\ell)$ (\myeqref{equation_avg_encoding}) smooths the averaged vector representations.
\end{enumerate*}
%
All encoding methods leveraging the generated labels $\ell$ outperform the baseline $E(x)$ using only the utterance, 
leading to ARI, NMI, and ACC gains between 5.12--19.23\%, 3.82--16.75\%, and 4.32--13.87\%, respectively.
This confirms our main hypothesis that abstractly summarizing utterances improves intent discovery.
Moreover, combining utterance and label encodings ($\phi_{\textsc{avg}}(x,\ell)$) further improves upon using the label alone (performing on par only for CLINC). 
Adding smoothing ($\phi_{\textsc{smooth}}(x,\ell)$) 
boosts performance even more.

\nonfirstparagraph{Inferring the number of smoothing neighbors:}
Smoothing requires selecting the number of neighbors $n'$.
Our proposed {\model} selects the value of $n' \in \lbrace$5, \ldots, 45$\rbrace$
that yields the highest silhouette score. 
To assess the effect of that chosen $n'$ value, we plot the ARI, NMI, and ACC scores for varying  $n'$ in \figref{fig:smoothing_neighbors}. 
We observe that the ARI, AMI, and ACC scores obtained with the automatically inferred $n'$ are nearly identical to the best achievable performance, demonstrating that the silhouette score is an effective heuristic for selecting a suitable number of smoothing neighbors.

\nonfirstparagraph{Random \vs nearest neighbor demonstrations:}
{\model} employs \textsc{kate} \cite{liu-etal-2022-makes} to select the $n$
ICL demonstrations most similar to $x$, \ie $\neighbors{n}{x}$, for generating $x$'s label (\secref{subsec:step2}).
To evaluate \textsc{kate}'s effectiveness for intent discovery, we present results for {\model} where 
$n$ (= 8) demonstrations are instead selected randomly.
\tabref{table:ablations_icl}
shows a substantial improvement of \textsc{kate} over the random selection method, where the latter only marginally outperforms {\model} \emph{without} any demonstrations (No ICL, $n=$~0). 
This follows the intuition that the LLM can pick a label from one of the $n$-NN instances, which likely shares an intent with the utterance to be labeled, thus effectively limiting label variation and improving clustering performance.

\nonfirstparagraph{Varying the number of ICL demonstrations:}
We generate labels
\begin{enumerate*}[(1)]
\item \emph{without} ICL, adopting the static prompt for generating the prototypical labels, without any demonstrations, and
\item \emph{with} ICL for varying numbers of demonstrations $n\in \lbrace$1, 2, \ldots, 16$\rbrace$.
\end{enumerate*}
\tabref{table:ablations_icl} shows that
\begin{enumerate*}[(i)]
\item using any number of demonstrations leads to superior performance compared to using no demonstrations (No ICL);
\item by varying small amounts of demonstrations ($n=$~1, 2, or 4) no 
significant differences are found; 
\item the best performance is achieved by using more demonstrations, \ie 8 or 16. Consistent with the results of \citet{min-etal-2022-rethinking, lyu2022z}, increasing $n$ from 8 to 16 does not result in further improvements, thus confirming that $n$\,=\,8 demonstrations is a good default value.
\end{enumerate*}

\begin{table}
\centering
\smaller
\setlength\tabcolsep{4pt} 
\begin{tabular}{lccc}
\toprule
& \multicolumn{3}{c}{StackOverflow} \\ 
\cmidrule(lr){2-4}
Method & ARI & NMI & ACC \\ \midrule
No ICL ($n=$ 0)& $\stdev{66.21}{0.13}$ & $\stdev{77.27}{0.04}$ & $\stdev{80.42}{0.13}$ \\
\midrule
\textsc{kate}, $n=$ 1
& $\stdev{68.91}{1.25}$ & $\stdev{79.11}{0.53}$ & $\stdev{83.09}{0.86}$ \\
\textsc{kate}, $n=$ 2
& $\stdev{68.88}{1.40}$ & $\stdev{79.06}{0.86}$ & $\stdev{82.67}{0.98}$ \\
\textsc{kate}, $n=$ 4
& $\stdev{69.97}{1.32}$ & $\stdev{79.76}{0.79}$ & $\stdev{82.94}{0.97}$ 
\\ 
\underline{\textsc{kate}, $n=$ 8}
& $\stdev{\underline{72.20}}{1.53}$ & $\stdev{\underline{81.26}}{0.93}$ & $\stdev{\underline{83.82}}{0.91}$ \\ 
\textsc{kate}, $n=$ 16
& $\stdev{72.49}{1.75}$ & $\stdev{82.07}{1.18}$ & $\stdev{83.50}{0.88}$ \\ 
\midrule 
random, $n=$ 8& $\stdev{66.80}{0.90}$ & $\stdev{78.72}{0.85}$ & $\stdev{81.37}{0.93}$\\
\midrule
 $K\times$2 ($n=$ 8)
 & $\stdev{71.43}{0.66}$ & $\stdev{80.76}{0.28}$ & $\stdev{83.51}{0.56}$ \\ 
\bottomrule
\end{tabular}
\caption{
ICL ablations. {\model} \underline{default} settings are $n=$ 8. The $K\times$~2 result uses twice the number of gold standard intents for the initial (Step~1, \secref{subsec:step1})
clustering (\ie 40 instead of 20 for StackOverflow).
}
\label{table:ablations_icl}
\end{table}

\nonfirstparagraph{Overestimating the number of prototypes:}
Following \citet{hadifar-etal-2019-self, zhang-etal-2021-supporting, Zhang_Xu_Lin_Lyu_2021, zhang-etal-2022-new}, {\model} assumes a known number $K$ of intents, both for the initial clustering (Step~1, retrieving prototypes, \secref{subsec:step1}) and for the final clustering (Step~4, recovering latent intents, \secref{subsec:step4}).
While $K$ can be \emph{estimated} from a subset of utterances, determining it exactly is difficult. Unlike MTP-CLNN \cite{zhang-etal-2022-new}, {\model} does not assume that the number of \emph{samples} of each latent intent is known. To probe the robustness of {\model}'s label generation to an incorrect number of prototypes, we conduct the initial {\kmeans} clustering with twice the gold number of intents. 
The $K\times$~2 row in \tabref{table:ablations_icl} shows
that this results in only a minor performance drop, indicating that {\model}'s label generation process is sufficiently robust to such overestimation. In fact, we hypothesize that having multiple prototypes representing the same intent is less harmful than an insufficient number or incorrectly selected prototypes that do not accurately represent each intent.

\section{Conclusions}
Unlike existing methods that \emph{train} unsupervised sentence encoders, our {\model} approach employs a \emph{frozen} pre-trained encoder since it increases the (dis)similarity of (un)related utterances in the textual space 
by abstractly summarizing utterances into ``labels''.
Our experiments demonstrate that {\model} substantially outperforms the current state-of-the-art in unsupervised intent discovery across multiple datasets (\ie Banking, StackOverflow, and our private Transport), and surpasses two recent semi-supervised methods on CLINC, despite not using any labeled intents at all.
Our findings suggest that our alternative strategy of abstractly summarizing utterances (using a general purpose LLM) is more effective than the dominant paradigm of training unsupervised encoders (specifically on dialogue data), and thus may open up new perspectives for novel intent discovery methods.
Since our generated labels provide a better measure of intent-relatedness,
we hypothesize that they could also enhance the performance of existing methods that train unsupervised encoders, \eg by
\begin{enumerate*}[(i)]
    \item reducing the number of false positive contrastive pairs for MTP-CLNN \cite{zhang-etal-2022-new}, or
    \item improving the purity of clusters induced by methods that iteratively cluster utterances and update the encoder with (self-)supervision from cluster assignments \cite{xie2016unsupervised, caron-etal, hadifar-etal-2019-self}.
\end{enumerate*}
To facilitate such follow-up work, we release our generated labels for the Banking, StackOverflow, and CLINC datasets.\footnote{\url{https://github.com/maarten-deraedt/IDAS-intent-discovery-with-abstract-summarization}.}


\section*{Limitations}
Our work is limited in the following senses. First, all presented results relied on the ground truth number of intents to initialize the number of clusters for conducting {\kmeans} to retrieve prototypes (\secref{subsec:step1}) and infer latent intents  (\secref{subsec:step4}).
In practice, however, the ground truth number of intents is unknown and needs to be estimated by examining a subset of utterances.
However, our ablations in \secref{subsec:ablations} investigated the impact of overestimating the number of ground truth intents by a factor of two, and found that {\model}'s performance did not degrade much.
While we did not explore this for the final {\kmeans} to infer latent intents, future work could investigate cluster algorithms that do not require the number of dialogue states as input, \eg DBSCAN \citep{ester1996density}, Mean shift \citep{comaniciu2002mean},
or Affinity propagation \citep{frey2007clustering}.

Second, we generated labels with the GPT-3 (175B) \texttt{text-davinci-003} model, which may be prohibitively expensive and slow to run for very large corpora.
In our initial experiments, we tried using smaller-sized models such as \texttt{text-curie-001}, \texttt{text-babbage-001}, and \texttt{text-ada-001}, as well as \texttt{Flan-T5-XL} \cite{flanXL}, but found that the generated labels were of lower quality compared to those of \texttt{text-davinci-003}.
In future work, it would thus be interesting to further explore how to more effectively exploit such smaller-sized and/or open-source language models.

\section*{Ethics Statement}

Since {\model} automatically recovers intents from utterances,
\eg those exchanged between users and support agents, any prejudices 
that may be present in these utterances may become apparent or even amplified in
intents inferred by our model, since clearly {\model} does not eliminate such prejudices.
Hence, when designing conversational systems based on such inferred intents, extra care should be taken to prevent them from carrying over to conversational systems deployed in the
wild.

Moreover, since {\model}'s label generation process relies on LLMs,
biases that exist in the data used to train these LLMs may be reinforced, leading to generated labels that may discriminate against or be harmful to certain demographics.

\section*{Acknowledgements}
This work was funded in part by Flanders Innovation \& Enterpreneurship (VLAIO), through Baekeland project-HBC.2019.2221 in collaboration with Sinch Chatlayer; and in part by the Flemish
government under the “Onderzoeksprogramma Artifici\"ele Intelligentie (AI) Vlaanderen” program.

\clearpage 

\bibliography{anthology,custom}
\bibliographystyle{acl_natbib}

\clearpage 

\appendix
\section{Appendix}
\label{sec:appendix}
In \secref{appendix:powerful}, we analyze how 
using a more powerful pre-trained sentence encoder affects the cluster performance of {\model}.
Additionally, we present and discuss the prompts in \secref{appendix:prompts}, and conduct a qualitative analysis of the generated labels produced by our {\model} approach in \secref{appendix:qualitative}. Finally, in \secref{appendix:computational}, we provide a brief overview of the implementation details of our experiments.

\begin{table*}[t]
\centering
\smaller
\begin{tabular}{l ccc ccc ccc ccc}
\toprule
& \multicolumn{3}{c}{Banking}  & \multicolumn{3}{c}{StackOverflow} & \multicolumn{3}{c}{CLINC} & \multicolumn{3}{c}{Average} \\ 
\cmidrule(lr){2-4} \cmidrule(lr){5-7} \cmidrule(lr){8-10} \cmidrule(lr){11-13}
Encoding & ARI & NMI & ACC  & ARI & NMI & ACC & ARI & NMI & ACC & ARI & NMI & ACC \\ \midrule
\multicolumn{13}{l}{MTP or \texttt{paraphrase-mpnet-base-v2}} \\ 
- $E(x)$ & 47.33 & 77.32 & 57.99 & 48.71 & 63.85 & 66.18  & 63.82 & 89.01 & 71.30 & 53.29 & 76.73 & 65.16 \\
- $E(\ell)$ & 52.45 & 81.14 & 62.31 & 67.94 & 80.60 & 80.05 & 75.01 & 93.04 & 81.27 & 65.13 & 84.93 & 74.54 \\ 
- $\phi_{\textsc{avg}}(x,\ell)$ & 54.47 & 82.35 & 63.25 & 69.20 & 80.76 & 81.29  & 75.65 & 93.33 & 81.04 & 66.44 & 85.48 & 75.19 \\
- $\phi_{\textsc{smooth}}(x,\ell)$ & 57.56 & 82.84 & 67.43 & 72.20  & 81.26 & 83.82  & 79.02 & 93.82 & 85.48 & 69.59 & 85.97 & 78.91\\ \midrule
\multicolumn{13}{l}{\texttt{all-mpnet-base-v2}} \\
- $E(x)$ & 54.09 & 81.29 & 64.27 & 57.69 & 72.40 & 71.72  & 69.24 & 91.05 & 76.04 & 60.34 & 81.58 & 70.68 \\
- $E(\ell)$ & 52.33 & 81.51 & 63.29 & 66.96 & 82.37 & 81.13  & 77.48 & 93.91 & 83.08 & 65.59 & 85.93 & 75.83 \\ 
- $\phi_{\textsc{avg}}(x,\ell)$ & 57.90 & 83.87 & 67.55 & 70.92 & 83.81 & 82.56  & 78.86 & 94.40 & 83.58 & 69.23 & 87.36 & 77.90 \\
- $\phi_{\textsc{smooth}}(x,\ell)$ & \textbf{59.88} & \textbf{84.13} & \textbf{70.07} & \textbf{78.27}  & \textbf{85.09} & \textbf{87.02}  & \textbf{82.26} & \textbf{94.93} & \textbf{87.80} & \textbf{73.47} & \textbf{88.05} & \textbf{81.84} \\
\bottomrule
\end{tabular}
\caption{\emph{Effect of using a more powerful sentence encoder.} The first four rows show the main results presented in \secref{subsec:main_results}, \ie with the MTP encoder for Banking and StackOverflow, and with \texttt{paraphrase-mpnet-base-v2} for CLINC.
The last four rows show the results of performing the final clustering (Step~4) with encoder \texttt{all-mpnet-base-v2}.}
\label{table:appendix_smpnet}
\end{table*}

\subsection{Effect of using a more powerful encoder}
\label{appendix:powerful}
Here, we assess the impact of using a more powerful frozen pre-trained encoder on the clustering performance of {\model}.
Specifically, we provide results of the four encoding strategies using the SBERT encoder \texttt{all-mpnet-base-v2} \cite{reimers-gurevych-2019-sentence} in \tabref{table:appendix_smpnet}.
The overall results, presented in the three rightmost columns as the average of the scores across the three datasets, show that each encoding strategy for \texttt{all-mpnet-base-v2} (bottom half of the table) consistently improves upon the corresponding results for the encoder used in our previous main results (as repeated here in the top rows). 
However, the label-only encoding strategy ($E(\ell)$) achieves similar results for different encoders, likely because the labels already are a short disambiguated version of their associated utterances. Conversely, the other three strategies that exploit the original utterances $x$ deliver substantially better results for \texttt{all-mpnet-base-v2}, as the advanced encoder can more effectively disambiguate utterances based on their latent intents, thus improving cluster performance.
Notably, using \texttt{all-mpnet-base-v2} for the smoothing strategy ($\phi_{\textsc{smooth}}(x,\ell)$) compared to using MTP (Banking, Stackoverflow) or \texttt{paraphrase-mpnet-base-v2} (CLINC), results in gains of +3.88\%, +2.08\%, and +2.93\% in ARI, NMI, and ACC, respectively. 

These results validate that employing more powerful pre-trained sentence encoders can further improve cluster performance out-of-the-box. It should be noted that, due to limitations in computation budget, we only replaced the encoder for Step~4 to induce intent clusters.
However, we anticipate that using \texttt{all-mpnet-base-v2} also for Steps 1--2 could result in additional improvements.

\subsection{Prompts}
\label{appendix:prompts}
\Figsref{figure:static_prompts}{figure:dynamic_prompts} present the static prompts used to generate prototypical labels in Step~2.1 (\secref{subsec:step2}) without demonstrations, as well as the ICL prompts for generating labels of non-prototypical utterances in Step~2.2 (\secref{subsec:step2}).
One advantage of instructing LLMs is the ability to specify additional information in the prompts. 
When clustering topic datasets, there typically is a general understanding of the broad topic according to which utterances should be partitioned, and this topic can be specified in the prompts used to instruct LLMs.
Since StackOverflow pertains to topics rather than intents, we adopted a more specific prototypical label generation prompt to instruct the LLMs to directly summarize the utterances based on the ``technology'' they refer to.
While this approach may not be effective for intent discovery (\ie a single conversational dataset can contain intents from multiple topics as well as non-topic intents), we speculate that it could be applied to other topic classification datasets, \eg News or Biomedical, where a prototypical prompt could instruct the LLM to identify the ``news category'' or ``medical drug'', ``disease'', 
\chris{etc.} 
We defer exploring {\model} for topic clustering beyond StackOverflow to future work.

\subsection{Qualitative Analysis}
\label{appendix:qualitative}
We conduct a qualitative analysis of {\model}'s generated labels.
\tabsref{appendix:samples_stackoverflow}{appendix:samples_clinc} show the generated labels for a subset of clusters induced in Step~4 for the corresponding StackOverflow, Banking, and CLINC datasets.
For each presented cluster, we report 
\begin{enumerate*}[(i)]
    \item the generated labels with their associated counts in that cluster, and
    \item the majority gold intent, \ie the most prevalent gold intent among utterances in that cluster, and the number of utterances within that cluster belonging to the majority gold intent.
\end{enumerate*}

\paragraph{Main findings:}
Overall, Tables \ref{appendix:samples_stackoverflow}--\ref{appendix:samples_clinc} reveal that there is little variation among generated labels within a specific cluster.
Specifically, for the majority of clusters, the most frequently occurring generated label has a notably higher count than the other generated labels, \eg the first row in  \tabref{appendix:samples_stackoverflow} shows that the label ``\emph{Magento}'' is generated for 47 out of 49 utterances in that cluster.
These findings further support our main hypothesis that abstract summarization increases the similarity in the input space of utterances with the same latent intent.
Given the low variation across generated labels within clusters, we hypothesize that our generated labels could also make clusters more easy to interpret compared to utterance-only clustering, thereby potentially reducing the time required for manually inspecting clusters in real-world settings.

\paragraph{Slightly specific labels:}
While most clusters clearly contain a single label that appears much more frequently than other labels, there are some clusters, \eg  \texttt{pto\_request}, \texttt{plug\_type}, \texttt{reminder\_update}, and \texttt{calories} for CLINC (\tabref{appendix:samples_clinc}), where this is not the case.
However, a closer examination of these clusters reveals that the labels still exhibit low variation since they share the same syntactic and lexical structure.
For instance, the \texttt{plug\_type} cluster's generated labels mostly follow the ``Plug Converter $\langle$noun adjunct$\rangle$'' pattern, with only the noun adjunct being specific to the utterance from which the label is generated. Note that for our intent discovery purpose, these slightly more specific labels do not negatively impact cluster performance, as long as there is a high overlap in syntactical and lexical structure among generated labels.

\paragraph{Overly general labels:}
Although some utterances are summarized into slightly more specific labels, others may be summarized into overly general labels. For instance, in the banking cluster \texttt{exchange\_via\_app} (\tabref{appendix:samples_banking}) the label ``\emph{Foreign currency exchange}'' appears 25 times.
However, 6 of those 25 utterances do not have \texttt{exchange\_via\_app} as their gold intent, despite having obtained the same generated label as those other 19 utterances that do.
This is due to the fact that generated labels corresponding to more high-level intents may be assigned to utterances that belong to different intents but share that common more high-level intent.
For instance, the utterances ``\emph{Can this app help me exchange currencies?}'' and ``\emph{I want to make a currency exchange to EU}'' have respective gold intents \texttt{exchange\_via\_app} and \texttt{fiat\_currency\_support}, yet both are summarized into a more high-level ``\emph{Foreign currency exchange}'' label.
In contrast to generated labels that are slightly too specific, overly general labels can adversely affect cluster performance, as they may incorrectly group together utterances  that belong to different intents despite sharing a common high-level intent.

\subsection{Implementation Details}
\label{appendix:computational}
For all presented experiments, the utterances are encoded (Steps 1, 3--4) on a 2.6 GHz 6-Core Intel Core i7 CPU, using a frozen pre-trained sentence encoder.
Similarly, both the initial and final {\kmeans} clustering to respectively retrieve prototypes (Step 1) and infer latent intents (Step 4), are conducted on CPU. We adopt the {\kmeans} implementation of \texttt{scikit-learn} \cite{pedregosa2011scikit}, with default parameter values, \ie using the algorithm of \citet{lloyd1982least} and \texttt{n\_init}\,=\,10.


\begin{figure*}
\centering
\begin{subfigure}{0.4\linewidth}
  \centering
  \begin{tikzpicture}
    \node[draw=mygreendark,rounded corners=2mm,text width=0.95\textwidth,inner sep=5mm,         font=\smaller, align=justify, line width=1pt] at (0,0) (box)
        {
        \vspace{0.2cm} \\ 
        Describe the banking question in a maximum of 5 words.\\
        \textbf{question:} \{prototype\} 
        \textbf{label:}};
        \node[yshift=-0.4cm] at (box.north) {Banking};
  \end{tikzpicture}
\end{subfigure}
\hspace{0.6cm}
\begin{subfigure}{0.4\linewidth}
  \centering
  \begin{tikzpicture}
    \node[draw=mygreendark,rounded corners=2mm,text width=0.95\textwidth,inner sep=5mm, font=\smaller, align=justify, line width=1pt] at (0,0) (box)
    {
    \vspace{0.2cm} \\
    Describe the transport question in a maximum of 5 words. \\
    \textbf{question:} \{prototype\}
    \textbf{label:}
    };
    \node[yshift=-0.4cm] at (box.north) {Transport};
  \end{tikzpicture}
\end{subfigure}

\begin{subfigure}{0.4\linewidth}
  \centering
  \begin{tikzpicture}
    \node[draw=mygreendark,rounded corners=2mm,text width=0.95\textwidth,inner sep=5mm, font=\smaller, align=justify, line width=1pt] at (0,0) (box)
    {
    \vspace{0.2cm} \\
    Describe the chatbot question in a maximum of 5 words.\\
    \textbf{question:} \{prototype\}
    \textbf{label:}};
\node[yshift=-0.4cm] at (box.north) {CLINC};
  \end{tikzpicture}
\end{subfigure}
\hspace{0.6cm}
\begin{subfigure}{0.4\linewidth}
  \centering
  \begin{tikzpicture}
    \node[draw=mygreendark,rounded corners=2mm,text width=0.95\textwidth,inner sep=5mm, font=\smaller, align=justify, line width=1pt] at (0,0) (box)
    {
    \vspace{0.2cm} \\
    Identify the technology in question.\\
    \textbf{question:} \{prototype\}
    \textbf{technology:} 
    \vspace{\baselineskip}
    };
    \node[yshift=-0.4cm] at (box.north) {StackOverflow};
  \end{tikzpicture}
\end{subfigure}
\caption{\emph{Static prototypical label generation prompts.} Note that since StackOverflow is a topic rather than an intent classification dataset, we adopt a slightly different prompt.}
\label{figure:static_prompts}
\end{figure*}

\begin{figure*}
\centering
\begin{subfigure}{0.4\linewidth}
  \centering
  \begin{tikzpicture}
    \node[draw=mybluedark,rounded corners=2mm,text width=0.9\textwidth,inner sep=5mm, font=\smaller, align=justify, line width=1pt] at (0,0) (box)
     {
        \vspace{0.2cm} \\
        Classify the transport question into one of the provided labels.\\
        \textbf{(1) question:} \{demonstration 1\}\\
        \textbf{(1) label:} \{label 1\} \\ 
        \textbf{(2) question:} \{demonstration 2\} \\
        \textbf{(2) label:} \{label 2\} \\
        \ldots \\
        \textbf{(8) question:} \{demonstration 8\} \\
        \textbf{(8) label:} \{label 8\} \\ \\
        \textbf{question:} \{input question\} \\
        \textbf{label:} \\ \\ \\ \\ 
        };
        \node[yshift=-0.4cm] at (box.north) {Transport};
  \end{tikzpicture}
\end{subfigure}
  \hspace{0.3cm}
\begin{subfigure}{0.4\linewidth}
  \centering
  \begin{tikzpicture}
    \node[draw=mybluedark,rounded corners=2mm,text width=0.9\textwidth,inner sep=5mm,         font=\smaller, align=justify, line width=1pt] at (0,0) (box)
        {
        \vspace{0.2cm} \\
        Classify the banking question into one of the provided labels.\\
        \textbf{(1) question:} My card is about to expire. How do I get a new one?\\
        \textbf{(1) label:} Get new card expiring \\
        \textbf{(2) question:} Can I get a spare card for someone else to use? \\
        \textbf{(2) label:} Additional card \\
        \ldots \\
        \textbf{(8) question:} What do I do when my card is about to expire? \\
        \textbf{(8) label:} Get new card expiring \\ \\
        \textbf{question:} Since my card is about to expire, I need a new one. \\
        \textbf{label:}
        };
        \node[yshift=-0.4cm] at (box.north) {Banking};
  \end{tikzpicture}
\end{subfigure}

\begin{subfigure}{0.4\linewidth}
  \centering
  \begin{tikzpicture}
    \node[draw=mybluedark,rounded corners=2mm,text width=0.9\textwidth,inner sep=5mm,         font=\smaller, align=justify, line width=1pt] at (0,0) (box)
        {
        \vspace{0.2cm} \\
        Classify the chatbot question into one of the provided labels.\\
        \textbf{(1) question:} Please tell me what kind of gas this car needs\\
        \textbf{(1) label:} Car gas type query \\
        \textbf{(2) question:} Is there a type of gas i need to use for this car \\
        \textbf{(2) label:} Car gas type query \\
        \ldots \\
        \textbf{(8) question:} how many miles per gallon do i get \\
        \textbf{(8) label:} Car gas mileage \\ \\
        \textbf{question:} What kind of gas will i need to put in this car \\
        \textbf{label:}
        };
        \node[yshift=-0.4cm] at (box.north) {CLINC};
  \end{tikzpicture}
\end{subfigure}
\hspace{0.3cm}
\begin{subfigure}{0.4\linewidth}
  \centering
  \begin{tikzpicture}
  \node[draw=mybluedark,rounded corners=2mm,text width=0.9\textwidth,inner sep=5mm,         font=\smaller, align=justify, line width=1pt] at (0,0) (box)
        {
        \vspace{0.2cm} \\
        Classify the question into one of the provided technologies.\\
        \textbf{(1) question:} When doing a tortoise svn merge, it includes a bunch of directories \ldots\\
        \textbf{(1) technology:} Subversion (SVN) \\
        \textbf{(2) question:} SVN how to resolve new tree conflicts when file is added on two branches \\
        \textbf{(2) technology:} Subversion (SVN) \\
        \ldots \\
        \textbf{(8) question:} how to put linq to sql in a separate project? \\
        \textbf{(8) technology:} LINQ to SQL \\ \\
        \textbf{question:} Using svn for general purpose backup. \\
        \textbf{technology:}
        };
        \node[yshift=-0.4cm] at (box.north) {StackOverflow};
  \end{tikzpicture}
\end{subfigure}
\caption{\emph{Prompts for non-prototypical label generation with ICL.}}
\label{figure:dynamic_prompts}
\end{figure*}

\begin{table*}[ht]
\smaller
\centering
\addtolength{\tabcolsep}{-2pt}   
\begin{tabularx}{\textwidth}{p{4.5cm}p{4cm}p{3.5cm}p{3.5cm}}
 \toprule
 \textbf{Majority gold topic} (\#\,$y_{\textsc{gold}}/|\mathcal{C}|$) & \multicolumn{3}{l}{\textbf{Generated labels} (\#\,$\ell$)} \\ \midrule
     \texttt{topic\_20} (49/49) & Magento (47) & Magento CodeIgniter (1) & Shipping Method (1) \\ \midrule
     \texttt{topic\_17} (44/45) & Drupal (35) & Drupal 6 (5) & Drupal 5 (1) \\
     & Drupal and Ruby on Rails (1) & Drupal Ubercart (1) & \textcolor{myreddark}{Web View (1)} \\ 
     & Drupal and Microsoft SQL Server and Microsoft IIS 7 (1) & & \\ \midrule
     \texttt{topic\_10} (43/49) & BASH scripting (30) & Shell Scripting (5) & Bash (Unix Shell) (2) \\ 
    & \textcolor{myreddark}{BASH scripting (2)}  & \textcolor{myreddark}{Scripting (1)} & Scripting (1) \\ 
     &  Shell scripting (1) & Pipe-separated files (1) & \textcolor{myreddark}{Readline (1)} \\
     & \textcolor{myreddark}{Scriptaculous (1)} & \textcolor{myreddark}{Shell Scripting (1)} & Bash scripting (1) \\ 
     & SSH scripting (1) & & \\ \midrule
     \texttt{topic\_6} (46/46) & Matlab (35) &  Matlab Octave (3) &  Matrix (1) \\ 
     &  MATLAB (1)  & MatLab Mathematica (1) &  MatLab (1) \\ 
     &  Matlab and C\# (1) &  N/A (1) & Image Processing (1)  \\ 
     &  Ezplot (Matlab plotting tool) (1) & &  \\ \midrule
     \texttt{topic\_19} (45/46) & Haskell (40) &  Haskell Cabal (1) &  General Programming (1) \\ 
     & Haskell HDBC (1) & \textcolor{myreddark}{ General Programming (1)} &  \\
     &  GHCi (Glasgow Haskell Compiler Interactive) (1)  & GHCI (Glasgow Haskell Compiler Interactive) (1) &  \\ \midrule
     \texttt{topic\_16} (42/45) & Qt (32) &  Qt C++ (2) & Qt (C++ library) (2) \\
     & Qt4 (1) & \textcolor{myreddark}{QT (1)} & QtScript (1) \\
     & Qt (C++) (1) & \textcolor{myreddark}{QuickTime (1)} &  IP Camera (1) \\
     &  Real Time Video Capture (1) & QT (1) & \textcolor{myreddark}{Quicksilver (1)} \\ \midrule
     \texttt{topic\_1} (45/48) & WordPress (38) & jQuery and cycle (1) & \textcolor{myreddark}{Drupal and WordPress (1)} \\ 
     &  \textcolor{myreddark}{Open Atrium (1)} & Disqus (1) & WordPress, PHP (1) \\ 
     & \textcolor{myreddark}{HTTP POST (1)} & Blogging (1) & WordPress and Django (1) \\ 
     & WordPress, RESTful, SOAP, InterWoven TeamSite (1)   & Commenting (1)  & \\ \midrule
     \texttt{topic\_{5}} (45/45) & Microsoft Excel (40) & Excel VBA (1) & Perl (1) \\ 
     & Microsoft Excel, Internet Information Services (IIS) (1) & Microsoft Excel, Visual Basic (1) & Google Earth (1) \\ \midrule
     \texttt{topic\_3} (47/53) & Subversion (SVN) (42) & \textcolor{myreddark}{File System (3)}
     & \textcolor{myreddark}{Subversion (SVN) (1)} \\ 
     & \textcolor{myreddark}{Apache web server and Subversion (SVN) (1)} & Subversion (SVN) and SharpSvn (1) &
     Version Control (1) \\ 
     & Subversion (SVN) and WebDAV (1) & Subversion (SVN) and Windows (1) & \\
     & Subversion (SVN) and Apache web server (1) & \textcolor{myreddark}{Concurrent Versions System (CVS) (1)} & \\ 
\bottomrule
\end{tabularx}
\addtolength{\tabcolsep}{2pt}  
\caption{Generated labels that occur in selected ${\model}$ clusters for StackOverflow, as well as the number of times \#\,$\ell$ each label $\ell$ occurs in corresponding cluster $\mathcal{C}$. The majority gold topic $y_{\textsc{gold}}$ of cluster $\mathcal{C}$ is the most prevalent gold topic among all utterances in $y_{\textsc{gold}}$, and \#\,$y_{\textsc{gold}}$ denotes the number of utterances in $\mathcal{C}$ with $y$\,=\,$y_{\textsc{gold}}$. Generated labels of utterances that have gold intents \textcolor{myreddark}{different} than $y_{\textsc{gold}}$ are highlighted in red. Since no descriptive topic names are provided for StackOverflow, we refer to them simply as numbered topics (\texttt{topic\_x})}.
\label{appendix:samples_stackoverflow}
\end{table*}

\begin{table*}[ht]
\smaller
\centering
\addtolength{\tabcolsep}{0pt}   
\begin{tabularx}{\textwidth}{p{4.85cm}p{4.35cm}p{5.5cm}}
 \toprule
    \textbf{Majority gold intent} (\#\,$y_{\textsc{gold}}/|\mathcal{C}|$) & \multicolumn{2}{l}{\textbf{Generated labels} (\#\,$\ell$)}\\ \midrule
    \texttt{lost\_or\_stolen\_phone} (38/38) & Lost phone banking app (37) & Switching phones banking app (1) \\ \midrule
    \texttt{atm\_support} (35/35) & ATM card acceptance (25) & Find nearest ATM (10) \\ \midrule
    \texttt{card\_acceptance} (24/27) & Card usage limits (24) & \textcolor{myreddark}{Card usage (3)} \\ \midrule 
    \texttt{virtual\_card\_not\_working} (31/33) & Virtual card not working (31) & \textcolor{myreddark}{Virtual card not received (2)}  \\ \midrule 
    \texttt{contactless\_not\_working} (37/39) &  Contactless banking issue (37) & \textcolor{myreddark}{Banking login issues (2)}  \\ \midrule
    \texttt{compromised\_card} (24/42) & Unauthorized card usage (24) & \textcolor{myreddark}{Unauthorized card usage (18)} \\ \midrule 
     \texttt{age\_limit} (39/39) & Age requirement for banking (30)
    & Opening an account for family members (9) \\ \midrule
    \texttt{terminate\_account} (40/41) & Close bank account (39) & Account closure advice (1)  \\
    & \textcolor{myreddark}{Change bank name (1)} \\ \midrule
    \texttt{card\_about\_to\_expire} (17/20) & Get new card expiring (17) & \textcolor{myreddark}{Get new card swallowed (3)} \\ 
    & \textcolor{myreddark}{Renew card banking (1)} & \\ \midrule 
    \texttt{card\_delivery\_estimate} (13/13) & Delivery time in US (9) & Delivery time request (2) \\
    & Delivery date selection (2) & \\ \midrule
    \texttt{country\_support} (17/17) & Banking countries operated in (14) & Banking locations (2) \\ 
    & Supported countries (1) & \\ \midrule
    \texttt{automatic\_topic} (27/27) & Automated top-up option (14) & Auto top-up location query (7) \\
    & Low balance top-up feature (5) & Auto top-up activation issue (1) \\ \midrule
    \texttt{receiving\_money} (14/18) & Banking - Salary Deposit (14) & \textcolor{myreddark}{Banking, Payment, Check (2)} \\ 
    & \textcolor{myreddark}{Banking - Types of Deposits (1)} & \textcolor{myreddark}{Banking, Deposit, Cheque (1)} \\ \midrule
    \texttt{receiving\_money} (10/19) & Configure salary in GBP (8) & \textcolor{myreddark}{Convert currency to GBP (2)} \\
    & Convert currency to GBP (1) & Deposit Money in GBP (1)  \\
    & \textcolor{myreddark}{Convert currency to AUD (6)} & \textcolor{myreddark}{Convert currency to AUD GBP (1)} \\ \midrule
    \texttt{apple\_pay\_or\_google\_pay} (40/40) & Top up with Google Pay (10) & Top up with Apple Pay (10) \\ 
    &  Apple Pay issue (10) & Top up with Apple Watch (8) \\
    & Cost of Apple Pay (1) & Set up Apple Pay (1) \\ \midrule
    \texttt{getting\_spare\_card} (22/25) 
    & Get second card banking (11) & Add card for family member (6) \\ 
    & Link existing bank card (4) & \textcolor{myreddark}{Link card to website (2)} \\
    &  Get spare card banking (1)  & \textcolor{myreddark}{Choose bank card (1)} \\ \midrule
    \texttt{visa\_or\_mastercard} (36/40) & Credit card offerings (19)& Credit card decision making (12)\\
    & Credit card application process (4) & \textcolor{myreddark}{Card payment acceptance (3)} \\
    & \textcolor{myreddark}{Credit card acceptance (1)} & Credit card eligibility (1) \\ \midrule
    \texttt{balance\_not\_updated\_after\_} & Cash deposit not posted (25) & Cash deposit pending query (6) \\ 
    \texttt{cheque\_or\_cash\_deposit} (36/38) & Cheque deposit processing time (1)  & Cash deposit not accepted (1) \\
    & Cash deposit flagged (1) & \textcolor{myreddark}{Cash deposit to account (1)} \\
    & \textcolor{myreddark}{Direct Deposit not posted (1)} &  \\ \midrule
    \texttt{exchange\_via\_app} (27/51) & Foreign currency exchange (19) & Currency exchange process (7) \\
    & Currency conversion (1) & \textcolor{myreddark}{Cryptocurrency exchange (7)} \\
    & \textcolor{myreddark}{Foreign currency exchange (6)} & \textcolor{myreddark}{Cross-border payments (1)} \\
    &  \textcolor{myreddark}{Receive payment in foreign currency (5)} &  \textcolor{myreddark}{Discounts for frequent currency exchange (5)} \\
\bottomrule
\end{tabularx}
\addtolength{\tabcolsep}{0pt}  
\caption{Generated labels that occur in selected ${\model}$ clusters for Banking, as well as the number of times \#\,$\ell$ each label $\ell$ occurs in corresponding cluster $\mathcal{C}$. The majority gold intent $y_{\textsc{gold}}$ of cluster $\mathcal{C}$ is the most prevalent gold intent among all utterances in $y_{\textsc{gold}}$, and \#\,$y_{\textsc{gold}}$ denotes the number of utterances in $\mathcal{C}$ with $y$=$y_{\textsc{gold}}$. Generated labels of utterances that have gold intents \textcolor{myreddark}{different} than $y_{\textsc{gold}}$ are highlighted in red.}
\label{appendix:samples_banking}
\end{table*}

\begin{table*}
\smaller
\centering
\addtolength{\tabcolsep}{-3pt}   
\begin{tabularx}{\textwidth}{p{4.5cm}p{5cm}p{5.75cm}}
 \toprule
    \textbf{Majority gold intent} (\#\,$y_{\textsc{gold}}/|\mathcal{C}|$) & \multicolumn{2}{l}{\textbf{Generated labels} (\#\,$\ell$)}\\ \midrule
    \texttt{find\_phone} (15/15) & 
    Locate Phone Request (15) & \\  \midrule
    \texttt{vaccines} (15/15) & 
    Travel Vaccination Needed (15) & \\ \midrule
    \texttt{exchange\_rate} (15/15) & Currency Exchange Rate (15) &  \\ \midrule
     \texttt{share\_location} (15/15) & Share Location Request (15) &  \\ \midrule 
    \texttt{international\_fees} (15/15) & International Transaction Fees (15) &  \\ \midrule
    \texttt{report\_fraud} (13/13) &
    Fraudulent Transaction Inquiry (11) & Report Fraudulent Activity (2)  \\ \midrule 
    \texttt{change\_speed} (15/15) &
    Speak slower please (8) & Speak faster please (7) \\ \midrule
    \texttt{tire\_pressure} (15/15) &
    Tire Air Pressure Query (14) & Tire air pressure query (1)  \\ \midrule
    \texttt{international\_visa} (15/16) &
    Need International Visa (15) & \textcolor{myreddark}{Intercontinental Meaning (1)} \\ \midrule 
    \texttt{pto\_request\_status} (13/17) & Vacation Request Status (12) &  Vacation request status (1) \\
    & \textcolor{myreddark}{Vacation Request Process (3)} & \textcolor{myreddark}{Vacation Request (1)} \\ \midrule
    \texttt{weather} (15/17) &  Weather forecast query (14) & Meteorological Data for Tallahassee (1) \\
    & \textcolor{myreddark}{AC Temperature Query (1)} & \textcolor{myreddark}{Set AC Temperature (1)} \\ \midrule
    \texttt{balance} (14/15) & Bank Account Balance (11) & Check Account Balance (2) \\ & bank account balance (1) & \textcolor{myreddark}{Bank Account Balance (1)} \\ \midrule
    \texttt{cancel\_reservation} (15/16) & Cancel restaurant reservation (8) & Cancel dinner reservation (4) \\ 
    & Cancel Reservations (1) & Call restaurant to cancel reservation (1) \\ 
    & Cancel reservation for Network (1) & \textcolor{myreddark}{Cancel Appointment (1)} \\ \midrule
    \texttt{pto\_request} (11/11) & PTO request for March (3) & PTO request for May (2) \\
    & PTO request for June (2) & PTO request for January (1) \\
    & PTO request for First to Ninth (1) & PTO request for January to February (1) \\
    & PTO request for July (1) & \\ \midrule
    \texttt{plug\_type} (15/15) &
    Plug Type Query (3) & Plug Converter Barcelona (2) 
    \\ & Plug Converter El Salvador (1) & Plug in electronics? (1) \\
    & Plug Converter Mexico (2)  & Plug Converter Thailand (1) \\
    & Plug Converter Denmark (1) & Plug Converter Israel (1) \\
    & Plug Converter Z (1) & Plug Converter Cairo (1) \\ 
    & Plug Converter Guam (1) & \\ \midrule 
    \texttt{reminder\_update} (14/28) & \textcolor{myreddark}{Ask Reminder List (9)} & 
    \textcolor{myreddark}{Remind of Forgotten Task (3)} \\ 
    & Set Reminder (3) & Set Reminder Later (2) \\ 
    &  \textcolor{myreddark}{Confirm Reminder Laundry (1)} & \textcolor{myreddark}{Set Reminder Later (1)}\\
    & Set Reminder Trash Out (1) & Set Reminder Dog Medicine (1) \\
    & Set Reminder Movie (1) & Set Reminder Pick Up Stan (1) \\
    & Set Reminder Bring Jacket (1) & Set Reminder Take Out Oven (1) \\ 
    & Set Reminder Conference (1) & Set Reminder Pay Bills (1) \\ 
    & Set Reminder Booking (1) & \\ \midrule
    \texttt{calories} (15/21) & Calorie content of apple (2) & Caloric value of cookie (1) \\
    &  Calorie content of peanut butter (1)  & Calorie content of fries (1) \\
    & Calorie content of Coke (1)  & Calorie content of whole cashews (1)  \\
    & Calorie content of bacon (1)  & Calorie content of cookie (1) \\
    & Calorie content of KitKat (1) & Calorie content of bagels (1) \\
    & Calorie content of Cheetos (1) & Calorie content of chocolate ice cream (2) \\
    & \textcolor{myreddark}{Nutrition Info for Brownies (1)} & \textcolor{myreddark}{Nutrition Facts for Cheerios (1)} \\
    & \textcolor{myreddark}{Health benefits of avocados (1)} & \textcolor{myreddark}{Health benefits of apples (1)} \\ 
    & \textcolor{myreddark}{Health benefits of chocolate (1)} & \textcolor{myreddark}{Nutrition Info for Lay's Potato Chips (1)} \\
    & Calorie content of Peanut Butter and Jelly Sandwich (1) & \\
\bottomrule
\end{tabularx}
\addtolength{\tabcolsep}{3pt}  
\caption{Generated labels that occur in selected ${\model}$ clusters for CLINC, as well as the number of times \#\,$\ell$ each label $\ell$ occurs in corresponding cluster $\mathcal{C}$. The majority gold intent $y_{\textsc{gold}}$ of cluster $\mathcal{C}$ is the most prevalent gold intent among all utterances in $y_{\textsc{gold}}$, and \#\,$y_{\textsc{gold}}$ denotes the number of utterances in $\mathcal{C}$ with $y$=$y_{\textsc{gold}}$. Generated labels of utterances that have gold intents \textcolor{myreddark}{different} than $y_{\textsc{gold}}$ are highlighted in red.}
\label{appendix:samples_clinc}
\end{table*}

\end{document}